\documentclass[runningheads]{llncs}
\usepackage{graphicx}

\usepackage[table]{xcolor}
\usepackage{tikz}
\usepackage{comment} 
\usepackage{amsmath,amssymb} 
\usepackage{color}
\usepackage{graphicx}
\usepackage{comment}
\usepackage{multirow}
\usepackage{hyperref}

\definecolor{whitebox}{RGB}{248, 164, 164}
\definecolor{graybox}{RGB}{248, 198, 198}
\definecolor{blackbox}{RGB}{248, 232, 232}
\definecolor{supervised}{RGB}{248, 115, 115}

\DeclareRobustCommand*{\ola}{\overleftarrow}
\newcommand{\otb}{OTB}
\newcommand{\dsi}{\mathcal{V}}
\newcommand{\stereo}{\mathcal{S}}
\newcommand{\imageL}{\mathcal{I}_L}
\newcommand{\imageR}{\mathcal{I}_R}
\newcommand{\dispL}{\mathcal{D}_L}
\newcommand{\dispR}{\mathcal{D}_R}

\newcommand{\mbce}{\mathcal{L}_\text{MBCE}}
\newcommand{\texture}{\mathcal{T}}
\newcommand{\agreement}{\mathcal{A}}
\newcommand{\uniqueness}{\mathcal{U}}

\newcommand{\eg}{e.g.}
\newcommand{\ie}{i.e.}

\begin{document}
\pagestyle{headings}
\mainmatter
\def\ECCVSubNumber{4771}  

\title{Self-adapting confidence estimation for stereo} 

\titlerunning{Self-adapting confidence estimation for stereo}
%
\author{
Matteo Poggi
\and
Filippo Aleotti
\and
Fabio Tosi
\and \\
Giulio Zaccaroni \and
Stefano Mattoccia
}
\authorrunning{M. Poggi et al.}
%
\institute{University of Bologna, Viale del Risorgimento 2, Bologna, Italy \\
}
\maketitle

\begin{abstract}
Estimating the confidence of disparity maps inferred by a stereo algorithm has become a very relevant task in the years, due to the increasing number of applications leveraging such cue.
Although self-supervised learning has recently spread across many computer vision tasks, it has been barely considered in the field of confidence estimation. 
In this paper, we propose a flexible and lightweight solution enabling self-adapting confidence estimation agnostic to the stereo algorithm or network. Our approach relies on the minimum information available in any stereo setup (\ie, the input stereo pair and the output disparity map) to learn an effective confidence measure. This strategy allows us not only a seamless integration with any stereo system, including consumer and industrial devices equipped with undisclosed stereo perception methods, but also, due to its self-adapting capability, for its out-of-the-box deployment in the field. Exhaustive experimental results with different standard datasets support our claims, showing how our solution is the first-ever enabling online learning of accurate confidence estimation for any stereo system and without any requirement for the end-user.
\end{abstract}

\begin{figure*}[t]
    \centering
    \renewcommand{\tabcolsep}{1pt}
    \begin{tabular}{ccccc}
        \includegraphics[width=0.19\textwidth]{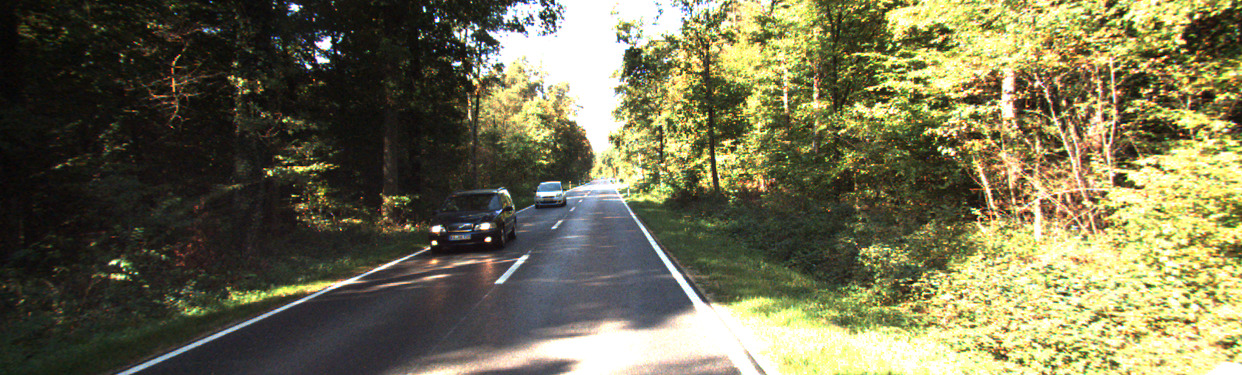} &
        \includegraphics[width=0.19\textwidth]{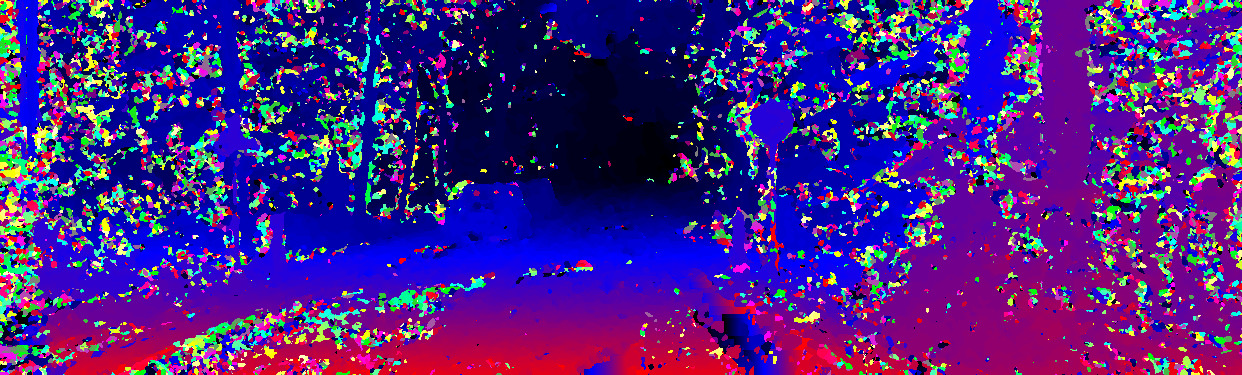} &
        \includegraphics[width=0.19\textwidth]{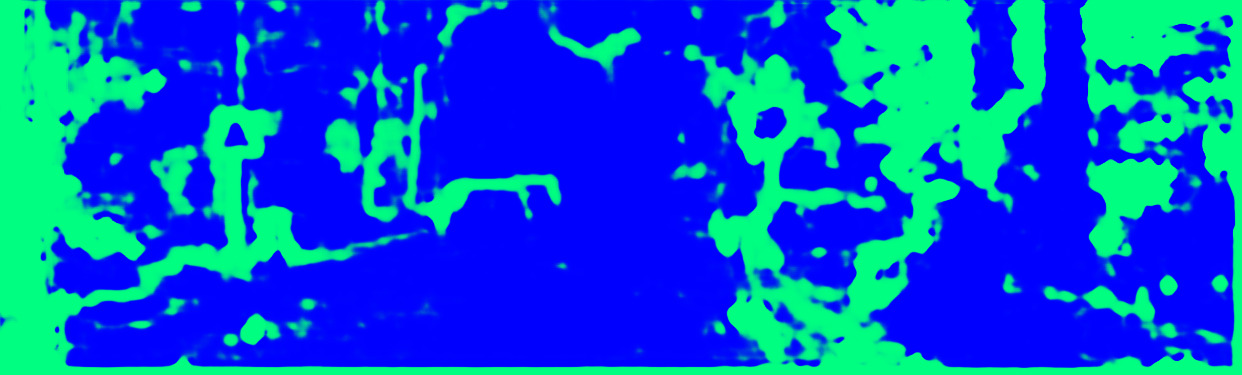} &
        \includegraphics[width=0.19\textwidth]{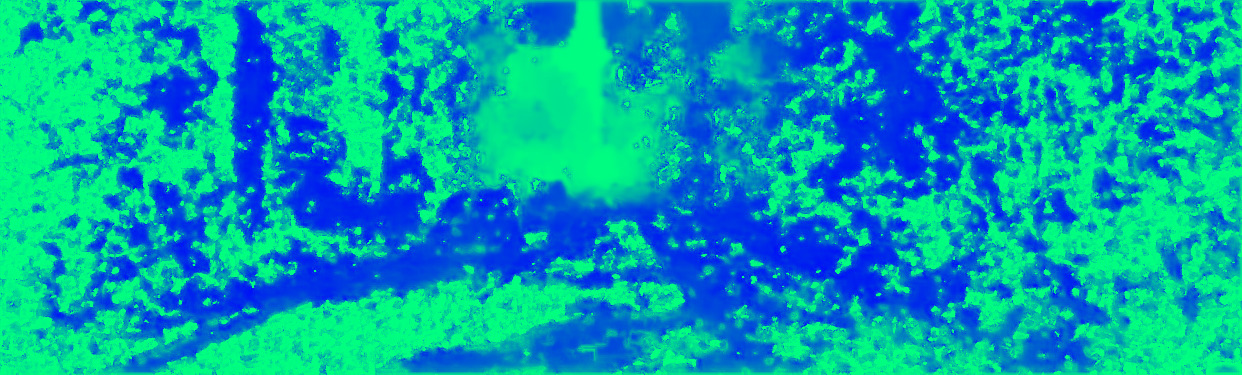} &
        \includegraphics[width=0.19\textwidth]{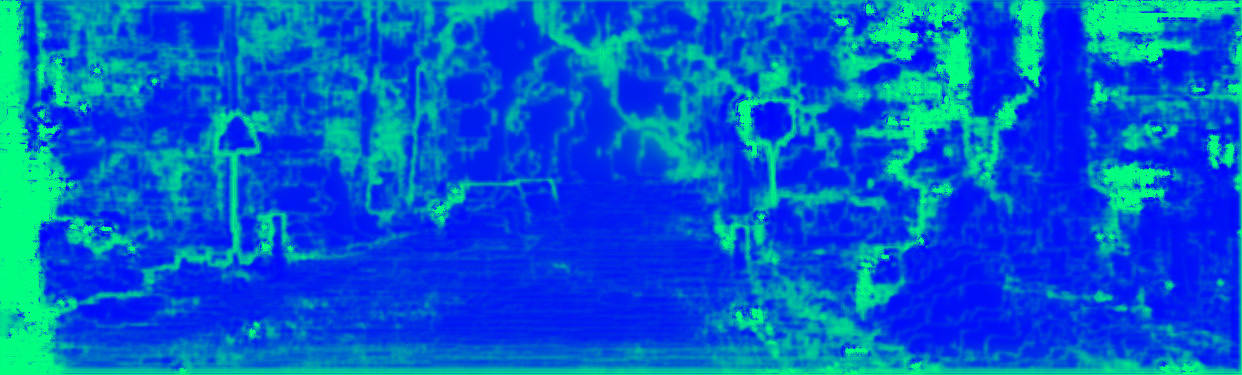}\\

        \includegraphics[width=0.19\textwidth]{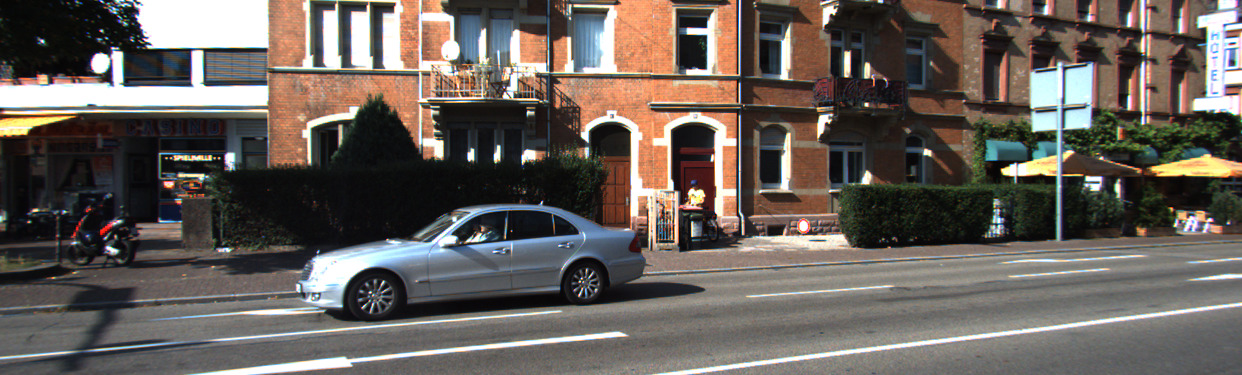} &
        \includegraphics[width=0.19\textwidth]{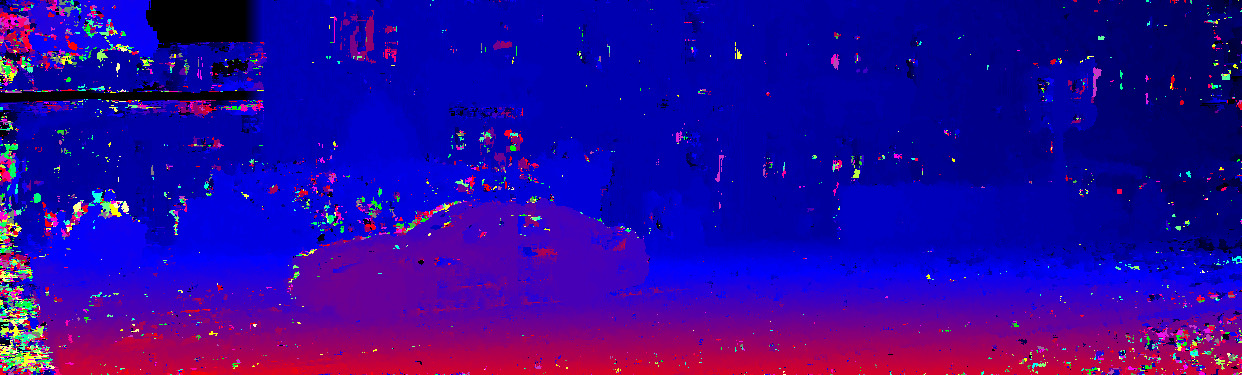} &
        \includegraphics[width=0.19\textwidth]{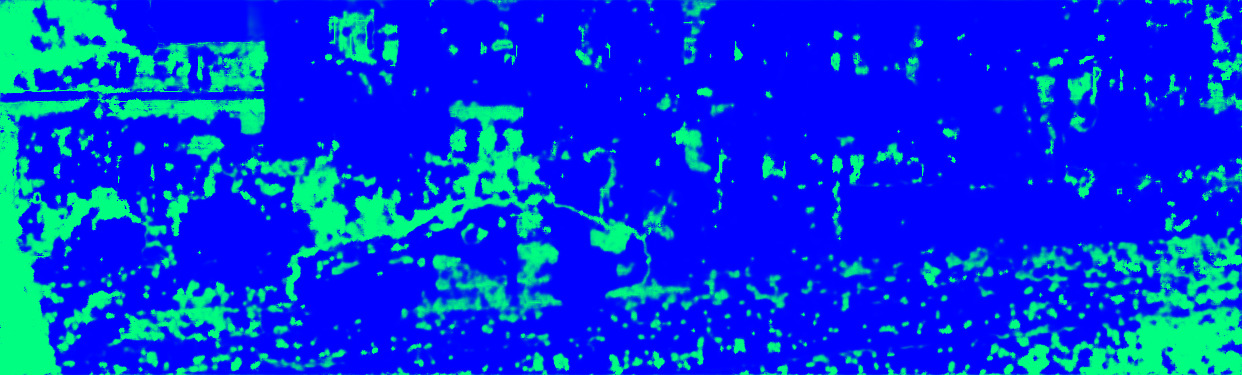} &
        \includegraphics[width=0.19\textwidth]{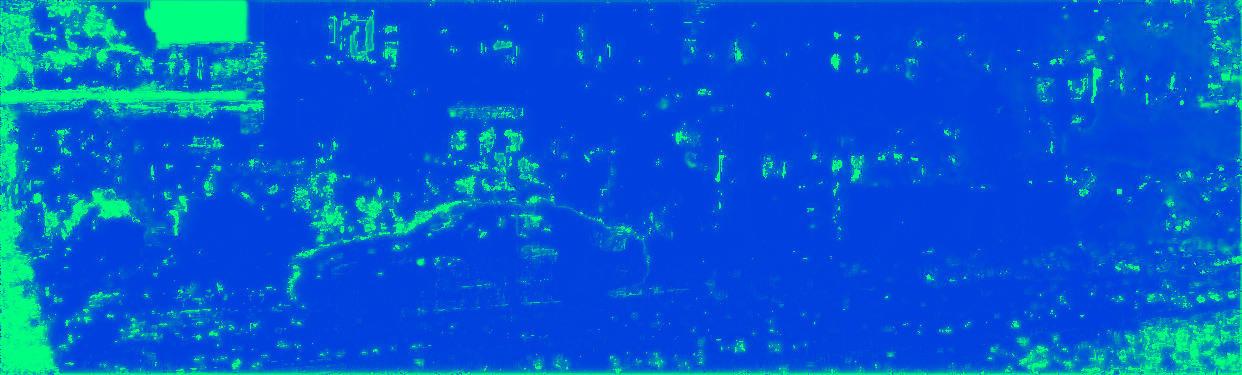} &
        \includegraphics[width=0.19\textwidth]{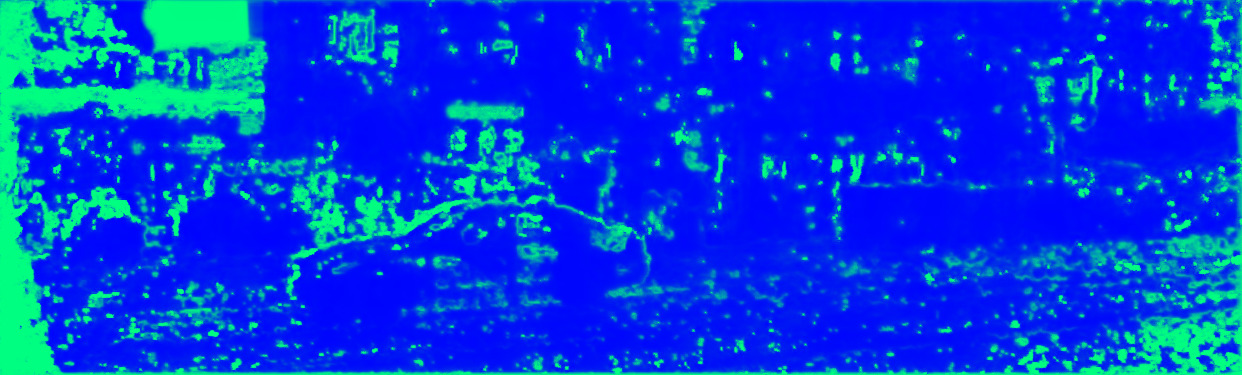}\\
        
        \includegraphics[width=0.19\textwidth]{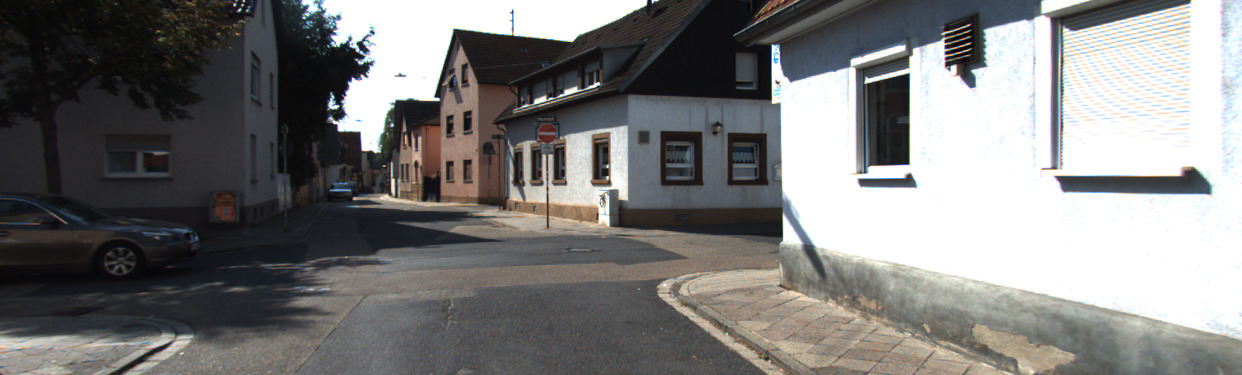} &
        \includegraphics[width=0.19\textwidth]{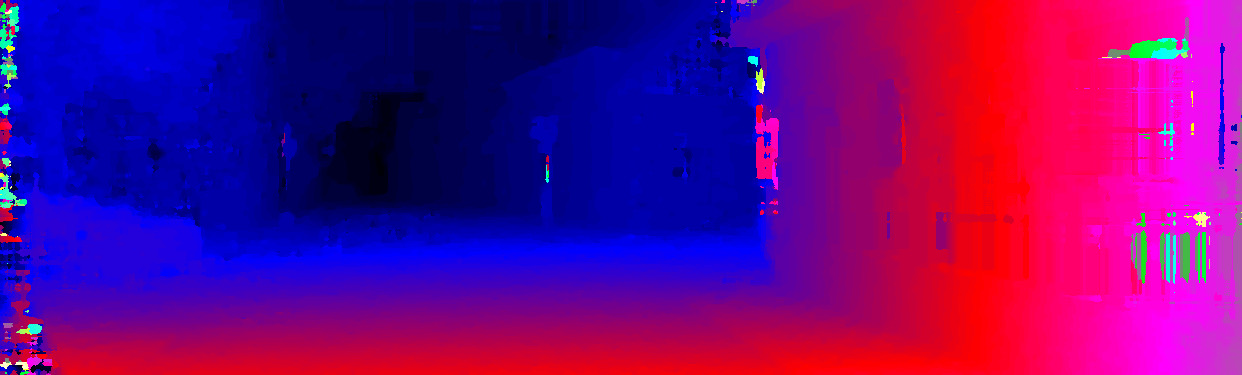} &
        \includegraphics[width=0.19\textwidth]{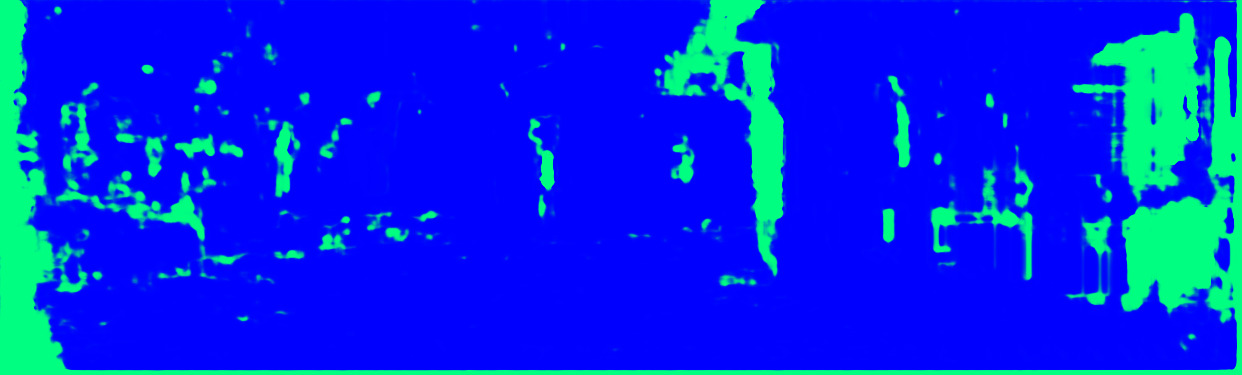} &
        \includegraphics[width=0.19\textwidth]{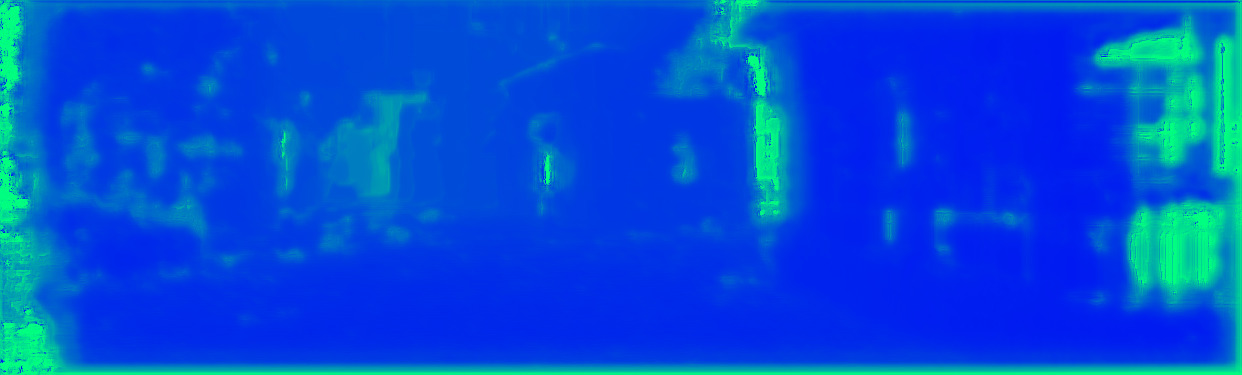} &
        \includegraphics[width=0.19\textwidth]{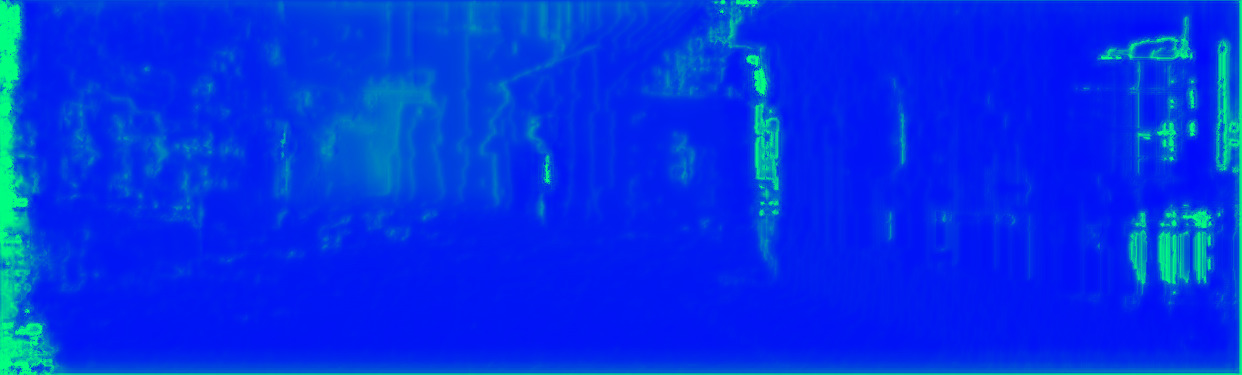}\\
        
        \includegraphics[width=0.19\textwidth]{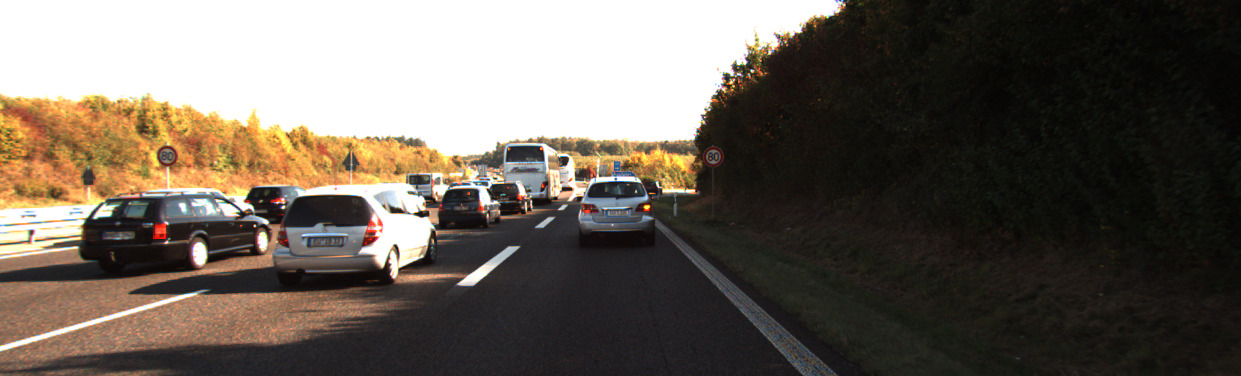} &
        \includegraphics[width=0.19\textwidth]{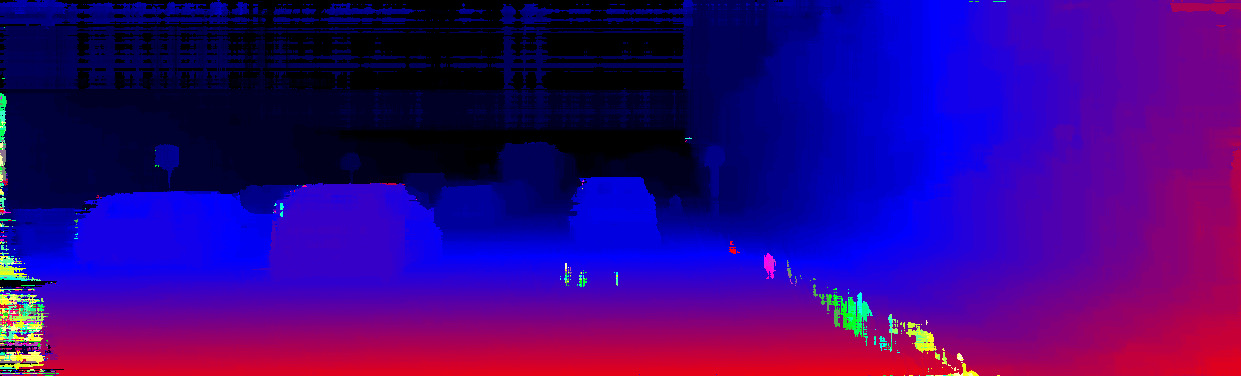} &
        \includegraphics[width=0.19\textwidth]{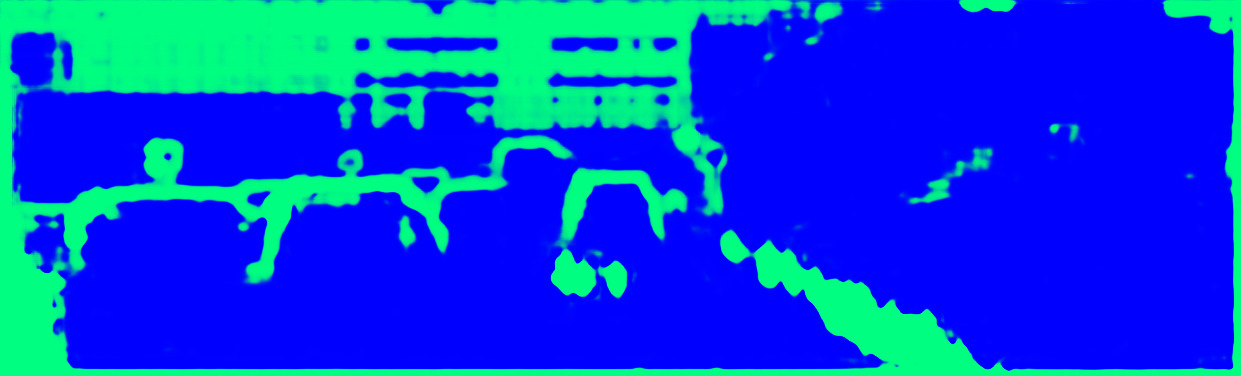} &
        \includegraphics[width=0.19\textwidth]{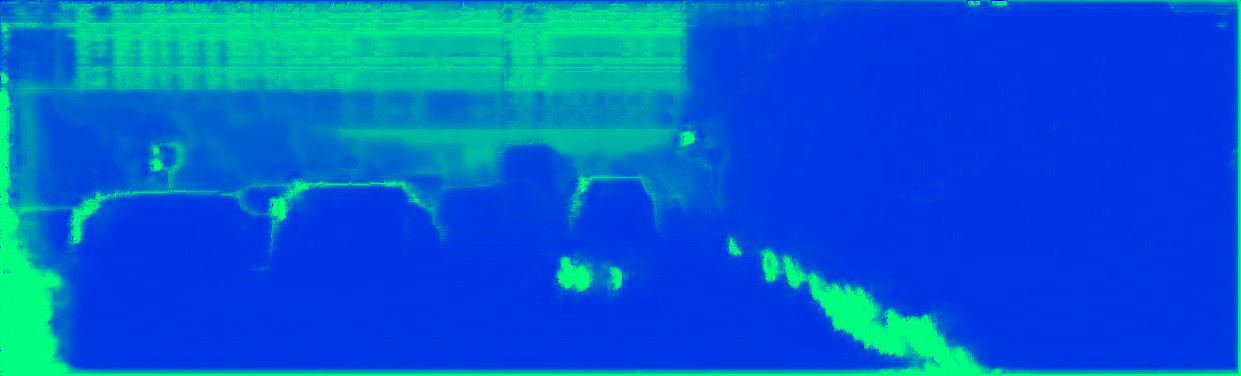} &
        \includegraphics[width=0.19\textwidth]{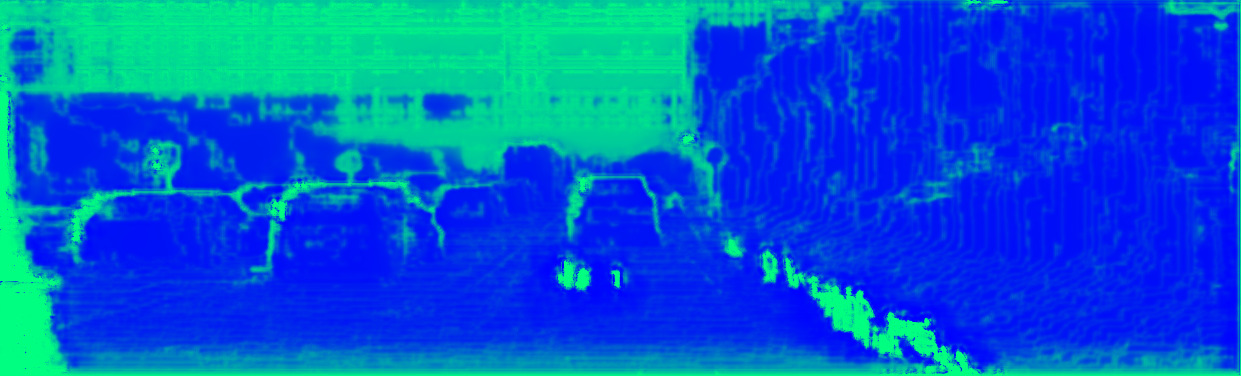}\\        
    \end{tabular}
    \caption{\textbf{Self-supervised confidence estimation.} From left, reference image, disparity from various algorithms and confidence estimated by self-supervised frameworks \cite{Tosi_2017_BMVC}, \cite{MOSTEGEL_CVPR_2016} and ours. From top to bottom: Census-CBCA, MCCNN-fst-CBCA, Census-SGM and MCCNN-fst-SGM. Color encoding details in the supplementary material.}
    \label{fig:abstract}
\end{figure*}

\section{Introduction}
Stereo is one of the most popular strategies to accurately perceive the 3D structure of the scene through two synchronized cameras and several algorithms, either hand-designed or based on deep neural networks, exist.
In many practical applications, alongside with disparity inference, confidence estimation is often performed as well. 
Purposely, a wide range of methods based either on hand-crafted measures \cite{hu2012quantitative} or \textit{learning-based} strategies \cite{poggi2017quantitative} have been proposed. 
Recent works \cite{tosi2018beyond,Kim_2019_CVPR,gul2019pixel} showed how state-of-the-art networks processing cues available from any stereo setup, \ie{} the input stereo pair and the output disparity map, are substantially equivalent to those processing the entire cost volume \cite{Kim_2019_CVPR}, 
further supporting the evidence that the disparity map itself contains sufficient clues to identify outliers as initially proposed in \cite{poggi2016learning,poggi2016bmvc}. Such a feature is highly desirable since it potentially paves the way for learning confidence estimation for any stereo camera even without any knowledge about the stereo algorithm/network deployed. This fact is very appealing since it frequently occurs with most industrial/off-the-shelf (\eg{} Stereolabs ZED 2) or consumer devices (\eg{} smartphones). Nonetheless, this opportunity was investigated only partially in the literature. 
Moreover, all these methods are strongly constrained to the need for ground truth depth labels acquired in the target domain. However, since achieving such labels is cumbersome and time-consuming, two self-supervised paradigms have been proposed in the literature \cite{MOSTEGEL_CVPR_2016,Tosi_2017_BMVC}. Although these methods proved that confidence estimation could be learned without needing active sensors, they have severe constraints. Individually, \cite{MOSTEGEL_CVPR_2016} requires static stereo sequences while \cite{Tosi_2017_BMVC} needs access to the \textit{cost volume}, rarely exposed in the case of off-the-shelf stereo sensors or not defined at all in most modern neural networks \cite{Mayer_2016_CVPR,Liang_2018_CVPR,Tonioni_2019_CVPR}. As a consequence, both are not thought to handle \textit{adaptation}, required to soften domain-shift issues.
Thus, a solution for out-of-the-box deployment of \textit{self-adapting} confidence estimation would be highly desirable for many practical applications.
A notable example concerns smartphone (\eg{} Apple iPhone) nowadays equipped with multiple cameras and undisclosed stereo algorithms/networks deployed for augmented reality or other applications in unpredictable environments.

Therefore in this paper, inspired by recent works performing continuous learning \cite{Tonioni_2019_CVPR,Casser_AAAI_2019} for depth estimation, we propose the first-ever solution for self-adaptation of a confidence measure unconstrained to the target stereo system. For this purpose, we deploy a novel loss function built upon cues available from the input stereo pair and the output disparity only, needing no additional information to learn/adapt to the sensed environment. Our solution is comparable, and often better, w.r.t known strategies requiring full access to the cost volume \cite{Tosi_2017_BMVC} or static scenes for training \cite{MOSTEGEL_CVPR_2016}, as shown in Fig. \ref{fig:abstract} on a variety of algorithms.

Extensive experimental results on KITTI, Middlebury 2014, ETH3D and DrivingStereo datasets support the following main claims of our novel confidence estimation paradigm: 1) competitive (often, better) with state-of-the-art when trained in a conventional, offline manner and tested on KITTI; 2) superior generalization capability on other datasets (\eg, Middlebury and ETH3D) compared to known self-supervised methods; 3) capable of online adaptation, outperforming competitors in unseen environments (\eg, DrivingStereo).

\section{Related work}

In this section, we review the literature concerning confidence measures and recent trends in stereo matching.

\textbf{Confidence measures for stereo.} Confidence measures have been, at first, reviewed and evaluated in \cite{hu2012quantitative} and, more recently, in \cite{poggi2017quantitative} highlighting that two broad categories exist: \textit{hand-made} and \textit{learned} measures. The former class consists of conventional method computed typically from cost volume analysis such as the ratio between two minima, as in PKR \cite{hirschmuller2002real}, or, as more recently proposed, determining local properties of the disparity map like the number of pixels with the same disparity hypothesis (DA \cite{poggi2016learning}).
Concerning learned measures \cite{poggi2017quantitative}, hand-made cues are usually combined and fed as input to a random forest classifier \cite{haeusler2013ensemble,spyropoulos2014learning,park2015leveraging,park2018learning,kim2017feature,poggi2016learning,poggi2020learning} or to a CNN \cite{seki2016patch,poggi2016bmvc,poggi2017learning,poggi2017even,fu2018learning,tosi2018beyond,Kim_2019_CVPR,gul2019pixel} appropriately trained deploying depth labels. 
Learned methods may require 1) full access to the cost volume to extract hand-made features \cite{haeusler2013ensemble,spyropoulos2014learning,park2015leveraging,park2018learning,kim2017feature,poggi2017even,poggi2017learning} or process the volume itself \cite{Kim_2019_CVPR,gul2019pixel,kim2019unified,Kim2020adversarial}, 2) disparity maps for both left and right viewpoint \cite{seki2016patch} or 3) only the input image and its corresponding disparity map \cite{poggi2016learning,poggi2020learning,poggi2016bmvc,fu2018learning,tosi2018beyond}. 
These three requirements translate into harder to softer constraints at deployment, most of them usually not met by off-the-shelf stereo cameras since exposing only the input stereo pair and the output disparity map to the user. 
Latest works \cite{Kim_2019_CVPR,gul2019pixel} showed that, although a CNN with access to the full cost volume can perform better than networks processing disparity and reference image only, the margin between the two approaches is small and in most cases negligible, at the cost of a much minor versatility of the former.

\textbf{Applications of confidence measures.} In addition to the traditional outliers filtering task, many higher-level applications exploit such cue for different purposes. Again, two main categories exist, acting inside a stereo algorithm or outside it.
Belonging to the former, Spyropoulos and Mordohai \cite{spyropoulos2014learning,spyropoulos2016correctness} estimate confidence and detect \textit{ground control points} to improve global optimization. Park and Yoon \cite{park2015leveraging,park2018learning} proposed a confidence-based modulation of the cost volume applied before SGM optimization, Poggi and Mattoccia \cite{poggi2016learning,poggi2020learning} reduced the streaking effects of the SGM \cite{hirschmuller2005accurate} stereo algorithm by using a weighted sum of the scanlines according to a confidence measure. Schonberger et al. \cite{schonberger2018sgm-forest} act similarly, fusing multiple scanlines of SGM using a random forest classifier. Seki and Pollefeys \cite{seki2016patch} changed P1 and P2 penalties of SGM dynamically according to the estimated confidence. 
Methods acting outside the stereo algorithms have been proposed for stereo algorithm fusion \cite{spyropoulos2015ensemble,poggi2016deep}, sensor fusion \cite{marin2016reliable,poggi2019confidence}, and unsupervised adaptation of deep models for stereo matching \cite{Tonioni_2017_ICCV,tonioni2019unsupervised}.

\textbf{Self-supervised confidence estimation.} Self-supervised learning has been barely investigated for confidence estimation. Mostegel et al. \cite{MOSTEGEL_CVPR_2016} leverage stereo videos looking at consistencies and contradictions between the different viewpoints of a static scene in order to obtain correct and wrong candidates from a given stereo algorithm.
Tosi et al. \cite{Tosi_2017_BMVC} instead rely on traditional confidence measures to obtain these two sets according to a consensus among them.

\textbf{Deep stereo and self-adaptation.} At first, CNNs have replaced single steps in the stereo pipeline \cite{scharstein2002taxonomy}, such as cost computation \cite{zbontar2016stereo,Chen_2015_ICCV,luo2016efficient}, rapidly converging towards end-to-end solutions estimating dense disparity maps by means of 2D \cite{Mayer_2016_CVPR,Liang_2018_CVPR,ilg2018occlusions,yin2019hierarchical,song2018edgestereo,yang2018segstereo,yin2019hierarchical} or 3D networks \cite{Kendall_2017_ICCV,Zhang2019GANet,Chang_2018_CVPR,Poggi_CVPR_2019,duggal2019deeppruner}.
The latest trend consists of training stereo networks from scratch using proxy labels \cite{Aleotti_ECCV_2020,Watson_ECCV_2020} or casting disparity estimation as a continuous learning problem. A first work in this latter direction is \cite{Zhong_ECCV_2018}, while more recent ones further moved in the direction of real-time continuous adaptation \cite{Tonioni_2019_CVPR} to new environments or meta-learning \cite{Tonioni_2019_learn2adapt}.

\section{Learning a confidence measure out-of-the-box}

This work aims at proposing a self-supervised paradigm suited for learning a confidence measure, unconstrained from the specific stereo method deployed and capable of self-adaptation. 
We first classify stereo systems into different categories according to the data they make available, and then we introduce a novel strategy compatible with all of them.

\subsection{Taxonomy of stereo matching systems}

In this section, we define three main broad categories of stereo matching solutions, each one characterized by different data made available during deployment.
From now on, we will refer to a generic rectified stereo pair as ($\imageL$,$\imageR$), respectively made of left and right images, and to a generic stereo algorithm or deep network as $\stereo$. In the remainder, to simplify notation, we omit $(x,y)$ coordinates if not strictly necessary.

\textbf{Black-box models.} Given any stereo algorithm processing a stereo pair ($\imageL$,$\imageR$), we define the output disparity map, computed assuming $\imageL$ as the \textbf{reference} image, as  
$\dispL=\stereo(\imageL,\imageR)$.
This image triplet is the minimum amount of data available out of any stereo method, and we define as \textbf{black-box} all the systems making available only these cues. Such systems are highly representative of off-the-shelf stereo cameras (\eg, Stereolabs ZED 2) or stereo methods implemented in consumer devices (\eg, Apple iPhones). They neither allow end-users to access the implementation nor provide explicit ways (APIs) to call for it.  For each ($\imageL$,$\imageR$) acquired in the field by the device, they provide the corresponding disparity map typically with undisclosed approaches based either on conventional stereo algorithms or deep networks.  
Hence, learning confidence measures for these systems is particularly challenging, yet appealing. 

\textbf{Gray-box models.} Although black-box systems provide cues available in any stereo system, when explicit calls to the algorithm APIs are exposed, additional cues can be retrieved. Hence, we define a second family of systems for which, although it is given no access to the algorithm implementation or its intermediate data, explicit calls to the method itself are possible (\eg{} stereo algorithms provided by pre-compiled libraries). 
Most deep stereo networks prevent the deployment of their internal representation since too abstract and substantially unintelligible, \eg{} 2D architectures \cite{Mayer_2016_CVPR,Liang_2018_CVPR,ilg2018occlusions,yin2019hierarchical,song2018edgestereo,yang2018segstereo}.
We define systems belonging to this class as \textbf{gray-box}, since multiple calls to $\stereo$ allow for retrieving additional cues. 
For instance, it is straightforward to compute the Left to Right Consistency (LRC) of the disparity maps, a popular strategy to obtain a confidence estimator, even if not explicitly provided by $\stereo${} itself in its original implementation. Given the possibility to call $\stereo${} two times, consistency checking can be performed analyzing $\dispL${} and a second disparity map, namely $\dispR$ obtained by assuming $\imageR${} as the reference images. Defining $\leftarrow{}$ the horizontal flipping operator, $\dispR${} is obtained as follows:
\begin{equation}
\label{eq:flip}
    \dispR = \ola{\stereo(\ola\imageR, \ola\imageL)}
\end{equation}
Once obtained $\dispR$, the consistency between the two can be checked as 
\begin{equation}
    \text{LRC} = | \dispL - \pi(\dispL, \dispR) | < \delta
\end{equation}
with $\pi(a,b)$ a sampling operator, collecting values at coordinate $a$ from $b$, and $\delta$ a threshold value (usually 1) above which $\dispL$ and $\dispR$ are considered inconsistent. Although less effective than other measures \cite{hu2012quantitative}, it comes at a lower price.

\textbf{White-box models.} Finally, if the implementation of $\stereo$ is accessible, additional cues can be sourced by processing intermediate data structures, if meaningful. The preferred one is the cost volume $\dsi$, containing matching costs $\dsi(x,y,d)$ for pixels at coordinates $(x,y)$ and any disparity hypothesis $d \in [0,d_{max}]$. This class of systems, referred to as \textbf{white-boxes}, enables computation of any confidence measure, either conventional \cite{hu2012quantitative} or learning-based \cite{poggi2017quantitative,Kim_2019_CVPR,gul2019pixel}. 
Popular traditional confidence measures obtained from $\dsi${} are the Peak-Ratio (PKR) and Left-Right Difference (LRD) defined, respectively, as 

\begin{equation}
    \text{PKR} = \frac{\dsi(d_{2m})}{\dsi(d_1)}
    \quad\text{and}\quad 
    \text{LRD} = \frac{\dsi(d_2)-\dsi(d_1)}{\dsi(d_1) - \min_{d} \dsi_R(x-d_1,y,d)}
\end{equation}
with $d_1$, $d_2$ and $d_{2m}$, respectively, the disparity hypotheses corresponding to the minimum cost, the second minimum and the second local minima \cite{hu2012quantitative}. Regarding LRD, given the cost volume $\dsi_R$ computed assuming $\imageR$ as the reference image, for any pixel $(x,y)$ costs are sampled at $(x-d_1,y)$.

\textbf{Motivations and challenges.}
Indeed, for the reasons outlined so far, black-box models represent the most challenging, yet general and appealing target when dealing with confidence estimation since their constraints prevent the deployment of most state-of-the-art measures \cite{Kim_2019_CVPR,gul2019pixel}, as well as self-supervised strategy existing in the literature \cite{MOSTEGEL_CVPR_2016,Tosi_2017_BMVC}. Hence, first and foremost, we aim at devising a general-purpose strategy enabling self-supervised confidence estimation in such constrained settings. As a notable consequence, this fact paves the way to tackle the same task even for state-of-the-art CNNs. 
Finally, having achieved this goal, out-of-the-box learning of confidence estimation with any stereo setup and self-adaptation in any environment is at hand.

\subsection{Self-supervision cues for black-box models}

In order to develop a self-supervised strategy suited for any stereo system, it is crucial to identify cues that are effective to source a robust supervision signal. According to the previous discussion, in the case of black-box models, we can rely on $(\imageL,\imageR)$ and $\dispL$ only. In this circumstance, although relevant information is not available compared to other models, we introduce three terms to obtain the desired self-supervised signal from the meagre cues available.

\textbf{Image reprojection error.} In recent literature, several works proved how the reprojection across the two viewpoints available in a rectified stereo pair could be a powerful source of supervision, either for monocular \cite{Godard_CVPR_2017,Poggi_IROS_2018,Godard_ICCV_2019} or stereo \cite{Zhang_ECCV_2018,Tonioni_2019_CVPR} depth estimation. Specifically, we can reproject $\imageR$ on the reference image coordinates as 
$
    \Tilde{\imageR} = \pi(\dispL, \imageR)
$
Then, the difference between $\imageL$ and warped right view $\Tilde{\imageR}$ appearance encodes how correct the reprojection is. To this aim, the most popular choice is a weighted sum between two terms, respectively SSIM \cite{wang2004image} and absolute difference.
\begin{equation}
    \Delta_{(\imageL,\Tilde{\imageR})} = \alpha \cdot (1-\text{SSIM}(\imageL,\Tilde{\imageR})) + (1-\alpha)|\imageL-\Tilde{\imageR}|
\end{equation}
with $\alpha$ usually tuned to 0.85 \cite{Godard_CVPR_2017}. The higher it is, the more likely $\dispL$ is wrong. 
By definition, matching pixels is particularly challenging in ambiguous regions, such as textureless portions of the image. To this aim, we first aim at detecting regions with rich texture, being more likely to be correctly estimated by $\stereo$, by comparing $\Delta$ computed between $(\imageL,\imageR)$ with the one after reprojection as
$
    \texture =  \Delta_{(\imageL,\imageR)} >  \Delta_{(\imageL,\Tilde{\imageR})} 
$.
In large ambiguous regions, $\Delta_{(\imageL,\imageR)}$ will result equal (or even minor) than the reprojection error \cite{Godard_ICCV_2019}, thus identifying pixels on which stereo is prone to errors. 

\textbf{Agreement among neighboring matches.} Since most regions of a disparity map should be smooth, variations in nearby pixels should be small except at depth boundaries. As highlighted in \cite{poggi2016learning,poggi2020learning}, $\dispL$ itself allows for the extraction of meaningful cues to assess the quality of disparity assignments. Purposely, we rely on the \textbf{disparity agreement} between neighbouring pixels, defined as

\begin{equation}
\text{DA} = \frac{\mathcal{H}_{N\times N}(d_1)}{N\times N}
\end{equation}
$\mathcal{H}_{N\times N}$ is a histogram encoding, for each pixel $(x,y)$, the number of neighbours in a $N \times N$ window having the same disparity $d$ (in case of subpixel precision, within 1 pixel). In the absence of depth discontinuities, the majority of pixels in the neighbourhood should share the same, or very similar, disparity hypothesis. Hence, we define a second criterion to identify reliable stereo correspondences as $\agreement=\text{DA} > 0.5$, assuming that more than half of the pixels in the neighbourhood share the same disparity. It is worth noting that this criterion is often not met in the presence of depth boundaries, even in case of correct disparities.

\textbf{Uniqueness constraint.} In an ideal frontal-parallel scene observed by a stereo camera in standard form, for each pixel in $\imageL$ exists at most one match in $\imageR$ and vice-versa. Leveraging this property, known as uniqueness, is particularly useful \cite{UC} to detect outliers in occluded regions and represents a reliable alternative to LRC and LRD measures, not usable when dealing with black-box models. Uniqueness Constraint (UC) is encoded as

\begin{equation}
    \text{UC} = \left[ x-\dispL(x,y) \right] \notin \bigcup_k \left[(x+k)-\dispL(x+k,y)\right]
\end{equation}
with $k \in [-\dispL(x,y),-1]\cup[1, d^*_{max}]$ and $d^*_{max}=d_{max}-\dispL(x,y)$. In other words, the uniqueness for any pixel in $\imageL$ holds if it does not collide in the target image with any other pixel, \ie, not matching the same pixel in $\imageR$ matched by any other. We exploit this property to define our third criterion as $\uniqueness=$UC. We conclude observing that, although effective at detecting mostly occlusions, the uniqueness constraint is often violated in the presence of slanted surfaces.

\begin{figure*}[t]
    \centering
    \includegraphics[width=1\textwidth]{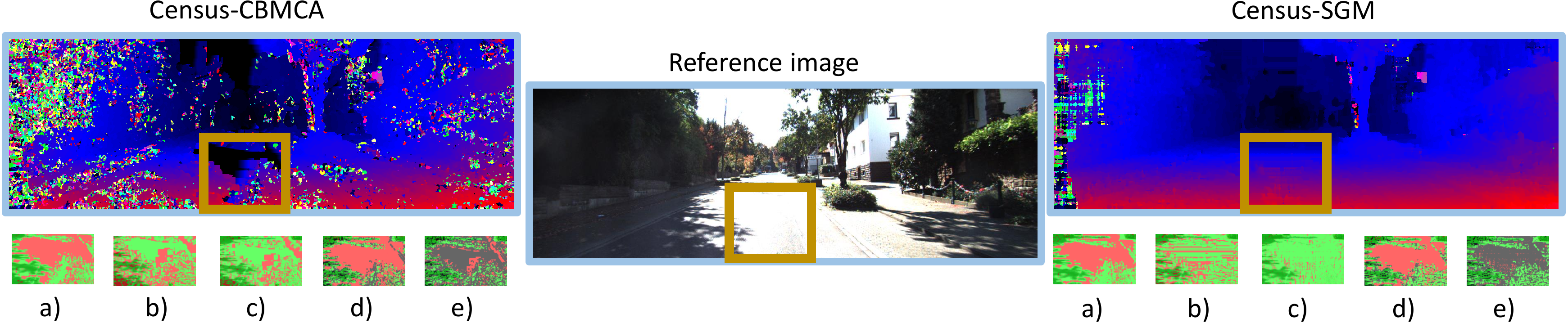}
    \caption{\textbf{Effects of different criteria.} Given the highlighted region, we show inliers (green) and outliers (red) guesses by using the following cues in multi-modal binary cross-entropy: a) $\texture^p,\texture^q$ b) $\agreement^p,\agreement^q$ c) $\uniqueness^p,\uniqueness^q$ d) $\texture^p,\agreement^p,\uniqueness^p,\texture^q$ e) $\texture^p,\agreement^p,\uniqueness^p,\texture^q,\agreement^q,\uniqueness^q$. For black pixels, the considered configuration gives no guesses. }
    \label{fig:labels}
\end{figure*}

\subsection{Multi-modal Binary Cross Entropy}

Given the three criteria outlined above, we revise the traditional binary cross entropy loss to take into account multiple label hypotheses. We refer to this variant as \textbf{Multi-modal Binary Cross Entropy} (MBCE), defined as

\begin{equation}
    \mathcal{L}_\text{MBCE} = - \left[ \left( \prod_{p \in \mathcal{P}}p \right) \cdot log(o) + \left( \prod_{q \in \mathcal{Q}} q \right) \cdot log(1-o)\right]
\end{equation}
with $o$ the output of the neural network $\in [0,1]$, \ie{} passed through a sigmoid activation, $\mathcal{P}$ and $\mathcal{Q}$ two sets of \textbf{proxy labels} derived respectively by a criterion being met or not. For instance, pixels satisfying the first criterion on image reprojection will have labels $\texture^p=1$, $\texture^q=0$ and vice versa when they do not. Unlike traditional binary cross entropy, where a single label $y$ and its counterpart $(1-y)$ are used, we define disjoint sets of proxies allowing for a flexible configuration of the loss function according to the three criteria described so far. For instance, by setting $\mathcal{P}=[\texture^p,\agreement^p]$ and $\mathcal{Q}=[\texture^q]$ we will train the network to detect good matches using image reprojection plus agreement and outliers using the former only. Adding elements to the sets $\mathcal{P}$ and $\mathcal{Q}$ reduces progressively the number of pixels considered correct or wrong, respectively. Fig. \ref{fig:labels} shows this, highlighting how combining multiple guesses as in d) and e) for some pixels no supervision is given when criteria do not match.
We will report the impact of this and the different configurations in a thorough ablation study.

\section{Experimental results}

In this section, we report an exhaustive evaluation to assess the effectiveness of our strategy, referred to as \textit{Out-of-The-Box} (\otb), by conducting three main experiments, respectively: 1) ablation study on the MBCE loss, 2) comparison with self-supervised approaches \cite{MOSTEGEL_CVPR_2016,Tosi_2017_BMVC} in a conventional offline training and 3) an evaluation concerning online adaptation of \otb{}. Code for this latter experiment is available at \url{http://github.com/mattpoggi/self-adapting-confidence}.

\subsection{Implementation details}

We now report all the details to understand and reproduce our experiments fully. 

\textbf{Evaluation Protocol.} To measure the effectiveness of the learned confidence measures, we compute the Area Under Curve (AUC) of the sparsification plots \cite{hu2012quantitative,poggi2017quantitative,tosi2018beyond,Kim_2019_CVPR}. Given a disparity map, pixels are sorted in increasing order of confidence and gradually removed (\eg, 5\% each time) from the disparity map. At each iteration, the error rate is computed over the sparse disparity map as the percentage of pixels having absolute error larger than $\tau$. Plotting the error rate results in a sparsification curve, whose AUC quantitatively assesses the confidence effectiveness (the lower, the better). Optimal AUC is obtained by sampling the pixels in decreasing order of absolute error or as in \cite{hu2012quantitative}. 

\textbf{Confidence networks.} Since the goal of this work is to define an effective self-supervised strategy suited for online learning rather than proposing a novel architecture, in our experiments, we test our proposal to train existing networks. Purposely, we consider three architectures: CCNN \cite{poggi2016bmvc}, ConfNet and LGC \cite{tosi2018beyond} to carry out our experiments because 1) they process only disparity map and reference image, thus are suited to all methods from white-box to black-box, 2) according to recent works \cite{Kim_2019_CVPR,gul2019pixel}, the most accurate one (LGC) is on par with state-of-the-art networks processing the cost volume and 3) the source code is fully available, conversely to \cite{Kim_2019_CVPR,gul2019pixel}. Moreover, in ConfNet we replaced deconvolutions with bilinear upsampling followed by $3\times3$ convolutions and process $\dispL$ only, significantly improving its performance and thus filling most of the gap with CCNN and LGC.
We defined a training schedule for each network, kept constant in all experiments. For CCNN, we use batches of 128 patches for 1M iterations, for ConfNet batches of single, 320$\times$1216 crops for 25K iterations, finally for LGC batches of 128 patches for 300K iterations, starting from pre-trained CCNN and ConfNet models. We trained all networks with SGD optimizer and a constant learning rate of 0.001. For patch-based methods, proxy signals are computed offline on the full resolution image.

\textbf{Datasets.} We consider five standard datasets: KITTI 2012 \cite{Geiger_CVPR_2012}, KITTI 2015 \cite{Menze2015CVPR}, Middlebury 2014\footnote{We use the quarter resolution split as in previous works \cite{poggi2017quantitative,tosi2018beyond,Kim_2019_CVPR}} \cite{scharstein2014high}, ETH3D \cite{schops2017multi} and DrivingStereo \cite{yang2019drivingstereo}, setting $\tau$ respectively to 3, 3, 1, 1 and 3. Being ground truth required to assess performance, we refer to the training set of such datasets. To train confidence estimation networks, we select the first 20 images from KITTI 2012 as in \cite{poggi2017quantitative,tosi2018beyond} for supervised training and the 400 images from the first 20 sequences of the KITTI 2012 multiview extension used in \cite{MOSTEGEL_CVPR_2016,Tosi_2017_BMVC} for self-supervised ones. 
To evaluate the trained confidence networks, we use the remaining 174 images from KITTI 2012 as the validation set and the totality of images available from KITTI 2015 for experiments on environments similar to the training set. Moreover, we also assess their generalization performance on the whole Middlebury 2014 and ETH3D datasets. In these experiments, only the KITTI 2012 images listed above are used for training, thus the networks are transferred without any fine-tuning.
Finally, to test self-adaptation peculiar of \otb{} we use a sequence from the DrivingStereo dataset, namely 2018-10-25-07-37, made of about 7K frames.

\textbf{Stereo algorithms.} Following the recent literature \cite{poggi2017quantitative,tosi2018beyond,Kim_2019_CVPR}, we evaluate the effectiveness of our strategy on a variety of stereo algorithms with different degrees of accuracy, in order to highlight how strong is our self-supervised paradigm in the presence of heterogenous disparity maps. We consider four main stereo algorithms deploying the code provided Zbontar and LeCun \cite{zbontar2016stereo} under different settings. Specifically: Census-CBCA, Census-SGM, MCCNN-fst-CBCA and MCCNN-fst-SGM. The first two rely on a census-based matching cost computation, respectively, optimized by a Cross Based Cost Aggregation (CBCA) strategy \cite{zhang2009cross} and SGM \cite{hirschmuller2005accurate}. The latter two replace the census-based matching costs with predictions obtained by MCCNN-fst, for which we use pre-trained weights on KITTI 2012, 2015 and Middlebury provided by the authors and tested on the same datasets. For ETH3D, Middlebury weights have been used. No post-processing is applied to any output.
Furthermore, to evaluate the impact of self-adaptation made possible by \otb{} with a real black-box method, we also consider two recent deep stereo network. We choose MADNet \cite{Tonioni_2019_CVPR} and GANet \cite{Zhang2019GANet}, both trained on synthetic images \cite{Mayer_2016_CVPR} and then fine-tuned with ground truth on KITTI 2015, because of the availability of trained model and its accuracy-speed trade-off. Since fine-tuned on KITTI, we conduct experiments with MADNet and GANet on DrivingStereo only.

\begin{table}[t]
    \centering
    \scalebox{0.7}{
    \renewcommand{\tabcolsep}{3pt}
    \begin{tabular}{c|c|c|c|c|c |r|r|r|r |r|r|r|r |r|r|r|r}
        \multicolumn{6}{c}{} & \multicolumn{12}{c}{KITTI 2012} \\
        \hline
         \multicolumn{6}{l|}{Match cost} & \multicolumn{2}{c|}{Census} & \multicolumn{2}{c|}{MCCNN-fst}& \multicolumn{2}{c|}{Census} & \multicolumn{2}{c|}{MCCNN-fst}& \multicolumn{2}{c|}{Census} & \multicolumn{2}{c}{MCCNN-fst} \\
        \hline
        \multicolumn{6}{l|}{Aggregation} & {CBCA} & {SGM} & {CBCA} & {SGM} & {CBCA} & {SGM} & {CBCA} & {SGM} & {CBCA} & {SGM} & {CBCA} & {SGM} \\
        \hline
        \hline
        
        \multicolumn{6}{l|}{$\Delta_{\mathcal{I}}(\imageL,\Tilde{\imageR})$} & 0.210 & 0.086 & 0.165 & 0.044 & 0.210 & 0.086 & 0.165 & 0.044 & 0.210 & 0.086 & 0.165 & 0.044 \\
        \multicolumn{6}{l|}{DA} & 0.112 & 0.047 & 0.063 & 0.023 & 0.112 & 0.047 & 0.063 & 0.023 & 0.112 & 0.047 & 0.063 & 0.023 \\
        \multicolumn{6}{l|}{UC} & 0.165 & 0.063 & 0.123 & 0.034 & 0.165 & 0.063 & 0.123 & 0.034 & 0.165 & 0.063 & 0.123 & 0.034 \\
        \hline\hline
        
        $\texture^p$ & $\agreement^p$ & $\uniqueness^p$& $\texture^q$ & $\agreement^q$ & $\uniqueness^q$& \multicolumn{4}{c|}{CCNN} & \multicolumn{4}{c|}{ConfNet} & \multicolumn{4}{c}{LGC} \\
         \hline
        $\checkmark$ & & & $\checkmark$ & & & 0.080 & 0.045 & 0.047 & 0.018 & 0.077 & 0.033 & 0.045 & 0.014 & 0.082 & 0.058 & 0.046 & 0.026 \\
        & $\checkmark$ & & &  $\checkmark$ & & 0.105 & 0.045 & 0.073 & 0.023 & 0.087 & 0.035 & 0.049 & 0.017 & 0.110 & 0.040 & 0.074 & 0.022 \\
        & & $\checkmark$ & & & $\checkmark$ & 0.111 & 0.035 & 0.087 & 0.022 & 0.101 & 0.038 & 0.065 & 0.020 & 0.114 & 0.035 & 0.077 & 0.020 \\
        \hline
        $\checkmark$ & $\checkmark$& & $\checkmark$ & & & 0.078 & 0.033 & 0.050 & 0.019 & 0.072 & 0.030 & 0.038 & 0.014 & 0.075 & 0.034 & 0.049 & 0.023 \\
        $\checkmark$ & $\checkmark$& & $\checkmark$ & $\checkmark$ & & 0.089 & 0.035 & 0.059 & 0.023 & 0.071 & 0.029 & 0.038 & 0.014 & 0.082 & 0.031 & 0.066 & 0.020 \\
        \hline
        $\checkmark$ & & $\checkmark$& $\checkmark$ & & & 0.072 & 0.038 & 0.053 & 0.019 & 0.074 & 0.029 & 0.040 & 0.013 & 0.070 & 0.036  & 0.042 & 0.016 \\
        $\checkmark$ & & $\checkmark$& $\checkmark$ & & $\checkmark$ & 0.088 & 0.032 & 0.075 & 0.020 & 0.076 & 0.030 & 0.041 & 0.013 & 0.084 & 0.031 & 0.071 & 0.017 \\ 
        \hline
        $\checkmark$ & $\checkmark$ & $\checkmark$ & $\checkmark$ & & & \bfseries\leavevmode\color{red} 0.068 & 0.034 & \bfseries\leavevmode\color{red} 0.046 & 0.018 & \bfseries\leavevmode\color{red} 0.070 & 0.029 & \bfseries\leavevmode\color{red} 0.037 & 0.013 & \bfseries\leavevmode\color{red} 0.068 & 0.032 & \bfseries\leavevmode\color{red} 0.041 & 0.016 \\
        $\checkmark$ & $\checkmark$& $\checkmark$ & $\checkmark$ & $\checkmark$ & $\checkmark$ & 0.085 & \bfseries\leavevmode\color{red} 0.029 & 0.057 & \bfseries\leavevmode\color{red} 0.017 & 0.071 & \bfseries\leavevmode\color{red} 0.028 & 0.038 & \bfseries\leavevmode\color{red} 0.012 & 0.081 & \bfseries\leavevmode\color{red} 0.028 & 0.050 & \bfseries\leavevmode\color{red} 0.015\\
        \hline
        \multicolumn{18}{c}{ } \\
        \multicolumn{6}{c}{} & \multicolumn{12}{c}{Middlebury} \\
        \hline
         \multicolumn{6}{l|}{Match cost} & \multicolumn{2}{c|}{Census} & \multicolumn{2}{c|}{MCCNN-fst}& \multicolumn{2}{c|}{Census} & \multicolumn{2}{c|}{MCCNN-fst}& \multicolumn{2}{c|}{Census} & \multicolumn{2}{c}{MCCNN-fst} \\
        \hline
        \multicolumn{6}{l|}{Aggregation} & {CBCA} & {SGM} & {CBCA} & {SGM} & {CBCA} & {SGM} & {CBCA} & {SGM} & {CBCA} & {SGM} & {CBCA} & {SGM} \\
        \hline
        \hline
        
        \multicolumn{6}{l|}{$\Delta_{I}(\imageL,\imageR)$} & 0.190 & 0.180 & 0.179 & 0.134 & 0.190 & 0.180 & 0.179 & 0.134 & 0.190 & 0.180 & 0.179 & 0.134 \\
        \multicolumn{6}{l|}{DA} & 0.161 & 0.168 & 0.099 & 0.087 & 0.161 & 0.168 & 0.099 & 0.087 & 0.161 & 0.168 & 0.099 & 0.087 \\
        \multicolumn{6}{l|}{$\uniqueness$} & 0.193 & 0.188 & 0.192 & 0.145 & 0.193 & 0.188 & 0.192 & 0.145 & 0.193 & 0.188 & 0.192 & 0.145 \\
        \hline\hline
        
         $\texture^p$ & $\agreement^p$ & $\uniqueness^p$& $\texture^q$ & $\agreement^q$ & $\uniqueness^q$& \multicolumn{4}{c|}{CCNN} & \multicolumn{4}{c|}{ConfNet} & \multicolumn{4}{c}{LGC} \\
        \hline
        $\checkmark$ & $\checkmark$ & $\checkmark$ & $\checkmark$ & & & \bfseries\leavevmode\color{red}0.116 & \bfseries\leavevmode\color{red}0.123 & \bfseries\leavevmode\color{red}0.087 & \bfseries\leavevmode\color{red}0.077 & \bfseries\leavevmode\color{red}0.133 & \bfseries\leavevmode\color{red}0.112 & \bfseries\leavevmode\color{red}0.087 & \bfseries\leavevmode\color{red}0.067 & \bfseries\leavevmode\color{red}0.127 & \bfseries\leavevmode\color{red}0.111 & \bfseries\leavevmode\color{red}0.090 & \bfseries\leavevmode\color{red}0.064 \\
        
        $\checkmark$ & $\checkmark$& $\checkmark$ & $\checkmark$ & $\checkmark$ & $\checkmark$ & 0.153 & 0.146 & 0.095 & 0.081 & 0.134 & 0.122 & 0.095 & 0.069 & 0.138  & 0.142 & 0.099  & 0.080 \\
        \hline
        
    \end{tabular}
    }
    \caption{\textbf{Ablation study on the proposed multi-modal binary cross entropy.} We report AUC scores for networks trained on KITTI 2012 (20 or 400 images) and tested on KITTI 2012 (174 images, top) and Middlebury (15 images, bottom).}
    \label{tab:ablation}
\end{table}

\textbf{Competitors.} We compare the proposed \otb{} strategy with existing methods proposed by Mostegel et al. \cite{MOSTEGEL_CVPR_2016} (named SELF) and by Tosi et al. \cite{Tosi_2017_BMVC} (named WILD). The former reasons about contradictions on observations from multiple viewpoints: given a stereo sequence framing a static scene with a moving camera, $\dispL$ and $\dispR$ are computed for each pair, registered and checked for inconsistencies. Since it requires both $\dispL$ and $\dispR$ disparity maps, SELF is suited only for systems belonging to gray-box and white-box categories.
Concerning WILD, it requires a pool of six confidence measures extracted from the cost volume to identify inliers and outliers according to heuristic thresholding on the measures. Since it requires access to the cost volume, WILD is suited for white-box algorithms only. 
In contrast, among other advantages discussed next, it worth stressing that our \otb{} approach is suited for black-box systems and agnostic to the scene content, in contrast to SELF that requires static scenes.

\begin{table}[t]
    \centering
    \scalebox{0.6}
    {
    \renewcommand{\tabcolsep}{12pt}
    \begin{tabular}{lll | rr | rr | rr | rr }
        \hline
        & & Match cost & \multicolumn{4}{c|}{Census} & \multicolumn{4}{c}{MCCNN-fst} \\
        \hline
        & & Aggregation & \multicolumn{2}{c|}{CBCA} & \multicolumn{2}{c|}{SGM} & \multicolumn{2}{c|}{CBCA} & \multicolumn{2}{c}{SGM} \\
        \hline
        & & KITTI split & 2012 & 2015 & 2012 & 2015 & 2012 & 2015 & 2012 & 2015 \\ 
        \hline
        & & Bad$\tau$ \% & 27.193 & 22.281 & 10.330 & 8.998 & 18.875 & 16.926 & 6.084 & 6.028 \\        
        \hline
        \hline
        
        \multirow{4}{*}{\rotatebox{90}{ Traditional }}
         & & \cellcolor{whitebox}LRD & 0.096 & 0.080 & 0.033 & 0.032 & 0.080 & 0.077 & 0.017 & 0.023 \\
         & & \cellcolor{whitebox}PKR & 0.106 & 0.089 & 0.028 & 0.029 & 0.065 & 0.062 & 0.010 & 0.017\\
        \cline{3-11}
        & & \cellcolor{graybox}LRC & 0.142 & 0.113 & 0.062 & 0.056 & 0.103 & 0.092 & 0.036 & 0.041 \\
        \cline{3-11}
         & & \cellcolor{blackbox}$\Delta_{(\imageL,\Tilde{\imageR})}$ & 0.210 & 0.175 & 0.086 & 0.079 & 0.165 & 0.150 & 0.044 & 0.041 \\
         & & \cellcolor{blackbox}DA & 0.112 & 0.090 & 0.047 & 0.046 & 0.063 & 0.059 & 0.023 & 0.028 \\
         & & \cellcolor{blackbox}UC & 0.165 & 0.131 & 0.063 & 0.058 & 0.123 & 0.111 & 0.034 & 0.037 \\
        
        \hline\hline
        
        \multirow{4}{*}{\rotatebox{90}{ CCNN }}
        & &\cellcolor{supervised}\textcolor{black}{Supervised} & 0.059 & 0.046 & 0.018 & 0.017 & 0.031 & 0.032 & 0.009 & 0.012 \\
        \cline{3-11}
        
         & & \cellcolor{whitebox}WILD \cite{Tosi_2017_BMVC} & 0.076 & 0.065 & \bfseries0.026 & \bfseries0.026 & 0.052 & 0.047 & \bfseries0.012 & \bfseries0.017 \\
         & &\cellcolor{graybox}{SELF \cite{MOSTEGEL_CVPR_2016}} & 0.076 & 0.065 & 0.047 & 0.046 & \bfseries0.038 & \bfseries0.041 & \bfseries0.012 & 0.018 \\
         & &\cellcolor{blackbox}\bfseries {\otb{} (Ours)} & \bfseries\leavevmode\color{red}0.068 & \bfseries\leavevmode\color{red}0.055 & 0.029 & 0.031 & 0.046 & 0.048 & 0.017 & 0.022\\
        
        \hline\hline
        
        \multirow{4}{*}{\rotatebox{90}{ ConfNet }}
        
        & &\cellcolor{supervised}\textcolor{black}{Supervised} & 0.061 & 0.049 & 0.017 & 0.016 & 0.033 & 0.034 & 0.006 & 0.010 \\
        \cline{3-11}
        
         & & \cellcolor{whitebox}WILD \cite{Tosi_2017_BMVC} & 0.089 & 0.067 & \bfseries\leavevmode\color{red}0.024 & \bfseries\leavevmode\color{red}0.020 & 0.054 & 0.050 & \bfseries\leavevmode\color{red}0.010 & \bfseries\leavevmode\color{red}0.016\\
         & &\cellcolor{graybox}{SELF \cite{MOSTEGEL_CVPR_2016}} & 0.075 & 0.066 & \bfseries\leavevmode\color{red}0.024 & 0.024 & 0.041 & 0.044 & 0.014 & \bfseries\leavevmode\color{red}0.016 \\
         & &\cellcolor{blackbox}\bfseries {\otb{} (Ours)} & \bfseries0.070 & \bfseries0.058 & 0.028 & 0.032 & \bfseries\leavevmode\color{red}0.037 & \bfseries\leavevmode\color{red}0.040 & 0.012 & 0.017\\

        \hline\hline 
         
        \multirow{4}{*}{\rotatebox{90}{ LGC }}
        
        & &\cellcolor{supervised}\textcolor{black}{Supervised} & 0.056 & 0.044 & 0.016 & 0.016 & 0.029 & 0.030 & 0.007 & 0.010 \\ 
        \cline{3-11}
        
         & & \cellcolor{whitebox}WILD \cite{Tosi_2017_BMVC} & 0.089 & 0.065 & \bfseries0.026 & \bfseries0.025 & 0.049 & 0.045 & \bfseries0.011 & \bfseries0.017\\
         & &\cellcolor{graybox}{SELF \cite{MOSTEGEL_CVPR_2016}} & 0.089 & 0.081 & \bfseries0.026 & 0.026 & 0.056 & 0.057 & 0.020 & 0.021 \\  
         & &\cellcolor{blackbox}\bfseries {\otb{} (Ours)} & \bfseries\leavevmode\color{red}0.068 & \bfseries\leavevmode\color{red}0.055 & 0.028 & 0.032 & \bfseries0.041 & \bfseries0.044 & 0.015 & 0.019 \\

        \hline\hline 
         & & Optimal & 0.047 & 0.034 & 0.008 & 0.008 & 0.024 & 0.022 & 0.003 & 0.005 \\    
        \hline
        
    \end{tabular}
    }
    \caption{\textbf{Evaluation on KITTI.} We report AUC scores for networks trained on KITTI 2012 (20 or 400 images) and tested on 2012 (174 images) and 2015 (200 images).}
    \label{tab:kitti}
\end{table}

\subsection{Ablation study}

At first, we study the impact of the different terms in the proposed self-supervised loss function. To this aim, on KITTI 2012 and as for other experiments, we train 9 variants of each network for each of the four stereo algorithms. Then, we evaluate confidences on the KITTI 2012 dataset and, without retraining, on Middlebury 2014. Table \ref{tab:ablation} collects the outcome of this evaluation, reporting  on top results on KITTI 2012 and, at the bottom, on Middlebury. We report as baselines the performance of $\Delta_{(\imageL,\Tilde{\imageR)}}$, DA and UC. DA is computed on $5\times5$ windows.

On KITTI (top of the table), we first report the results achieved by training the three networks selecting only one of the three cues used to distinguish between correct and wrong matches, \ie{} $[\texture^p,\texture^q]$, $[\agreement^p,\agreement^q]$ and $[\uniqueness^p,\uniqueness^q]$ configurations. We can notice that each of them outperforms the performance of the corresponding baseline used for supervision. This trend occurs on all the algorithms and for each network, showing the surprisingly robust capacity of the networks to learn how to estimate confidence better than a noisy supervision signal used for training. In general, the models trained on $[\texture^p,\texture^q]$ outperforms the others, except rare cases (\ie{} CCNN and LGC on Census-SGM, outperformed by $[\uniqueness^p,\uniqueness^q]$ setting).  
Although effective at detecting textureless and ambiguous regions, the reprojection fails at filtering outliers due to slanted surfaces and occlusions. Thus, we incrementally add a single criterion, \ie{} $\agreement^p$ or $\uniqueness^p$ to filter out false positives obtained by $[\texture^p,\texture^q]$ configuration. We incrementally add, on another configuration, the corresponding negative criterion to remove pixels wrongly categorized as outliers by $\texture^q$. In most cases, adding a single criterion to $\mathcal{P}$ is beneficial, while we can notice how introducing negative criteria degrades the performance on CBCA algorithms. This occurs because adding $\agreement^q$ or $\uniqueness^q$ makes textureless regions no longer labelled as outliers, as shown in Fig. \ref{fig:labels} left comparing patches d) and e).
Finally, adding both $\agreement^p$ and $\uniqueness^p$ produces the best overall results for CBCA methods. By introducing $\agreement^q$ and $\uniqueness^q$ too we obtain better results only on SGM methods, since much more accurate than CBCA ones and thus more false outliers are introduced if $\agreement^q$ and $\uniqueness^q$ are not used, as shown in Fig. \ref{fig:labels} right, comparing d) and e). 
On the other hand, by testing the best configurations on Middlebury 2014, enabling all the positive criteria and only $\texture^q$ for negative allows for better generalization to unseen environments.

\subsection{Comparison with offline methods}

Having found the best configuration for the $\mbce$ loss, we compare our supervision paradigm with known self-supervised approaches \cite{MOSTEGEL_CVPR_2016,Tosi_2017_BMVC}. In our experiments, we obtain proxy labels for SELF and WILD using the code provided by the respective authors. We collect the outcome of these experiments in Tables \ref{tab:kitti} and \ref{tab:middeth}. We label with different colors methods ranging from \textcolor{red}{\textbf{stronger}} constraints (need for ground truth) to \textcolor{whitebox}{\textbf{weaker}} (ours). For each architecture, stereo algorithm and evaluation set triplet we label in \textbf{bold} the best self-supervision approach, while in \textbf{\textcolor{red}{red}} the couple architecture/self-supervision on an entire evaluation set. 

\begin{table}[t]
    \centering
    \scalebox{0.6}
    {
    \renewcommand{\tabcolsep}{12pt}
    \begin{tabular}{lll | rr | rr | rr | rr }
        \hline
        & & Match cost & \multicolumn{4}{c|}{Census} & \multicolumn{4}{c}{MCCNN-fst} \\
        \hline
        & & Aggregation & \multicolumn{2}{c|}{CBCA} & \multicolumn{2}{c|}{SGM} & \multicolumn{2}{c|}{CBCA} & \multicolumn{2}{c}{SGM} \\
        \hline
        & & Dataset & Midd & ETH & Midd & ETH & Midd & ETH & Midd & ETH \\ 
        \hline
        & & Bad1 \% & 28.701 & 21.270 & 26.682 & 15.471 & 29.799 & 34.279 & 21.799 & 12.594 \\
        \hline
        \hline
        
        \multirow{4}{*}{\rotatebox{90}{ Traditional }}
         & & \cellcolor{whitebox}LRD & 0.117 & 0.082 & 0.113 & 0.059 & 0.107 & 0.185 & 0.075 & 0.051 \\
         & & \cellcolor{whitebox}PKR & 0.124 & 0.086 & 0.112 & 0.056 & 0.095 & 0.181 & 0.059 & 0.042 \\
        \cline{3-11}
        & & \cellcolor{graybox}LRC & 0.189 & 0.135 & 0.197 & 0.114 & 0.188 & 0.239 & 0.149 & 0.091 \\
        \cline{3-11}
         & & \cellcolor{blackbox}$\Delta_{(\imageL,\Tilde{\imageR})}$ & 0.190 & 0.162 & 0.180 & 0.119 & 0.179 & 0.257 & 0.134 & 0.097 \\
         & & \cellcolor{blackbox}DA & 0.161 & 0.119 & 0.168 & 0.093 & 0.099 & 0.159 & 0.087 & 0.047 \\
         & & \cellcolor{blackbox}UC & 0.193 & 0.148 & 0.188 & 0.114 & 0.192 & 0.264 & 0.145 & 0.096 \\
        
        \hline\hline
        
        \multirow{4}{*}{\rotatebox{90}{ CCNN }}
        
         & &\cellcolor{supervised}\textcolor{black}{Supervised} & 0.110 & 0.096 & 0.118 & 0.076 & 0.079 & 0.138 & 0.068 & 0.046 \\
         \cline{3-11}        
        
         & & \cellcolor{whitebox}WILD \cite{Tosi_2017_BMVC} & 0.136 & 0.114 & 0.140 & 0.086 & 0.095 & 0.154 & 0.081 & 0.046 \\
         & &\cellcolor{graybox}{SELF \cite{MOSTEGEL_CVPR_2016}} & 0.163 & 0.174 & 0.217 & 0.174 & 0.090 & 0.147 & 0.081 & 0.076 \\
         & &\cellcolor{blackbox}\bfseries {\otb{} (Ours)} & \bfseries\leavevmode\color{red}0.116 & \bfseries\leavevmode\color{red}0.084 & \bfseries0.123 & \bfseries0.070 & \bfseries\leavevmode\color{red}0.087 & \bfseries\leavevmode\color{red}0.137 & \bfseries0.077 & \bfseries0.042 \\

        \hline\hline
        
        \multirow{4}{*}{\rotatebox{90}{ ConfNet }}
        
         & &\cellcolor{supervised}\textcolor{black}{Supervised} & 0.121 & 0.086 & 0.104 & 0.063 & 0.086 & 0.138 & 0.062 & 0.036 \\
         \cline{3-11}        
        
         & & \cellcolor{whitebox}WILD \cite{Tosi_2017_BMVC} & \bfseries0.122 & 0.101 & 0.117 & \bfseries0.063 & 0.091 & 0.160 & 0.073 & 0.037 \\
         & &\cellcolor{graybox}{SELF \cite{MOSTEGEL_CVPR_2016}} & 0.154 & 0.120 & 0.121 & 0.067 & 0.096 & 0.172 & 0.084 & 0.048 \\
         & &\cellcolor{blackbox}\bfseries {\otb{} (Ours)} & 0.133 & \bfseries0.093 & \bfseries0.112 & 0.067 & \bfseries\leavevmode\color{red}0.087 & \bfseries0.138 & \bfseries0.067 & \bfseries\leavevmode\color{red}0.035 \\
         
        \hline\hline 
         
        \multirow{4}{*}{\rotatebox{90}{ LGC }}
        
         & &\cellcolor{supervised}\textcolor{black}{Supervised} & 0.111 & 0.080 & 0.111 & 0.061 & 0.083 & 0.136 & 0.065 & 0.040 \\ 
         \cline{3-11}
        
         & & \cellcolor{whitebox}WILD \cite{Tosi_2017_BMVC} & 0.136 & 0.104 & 0.133 & 0.082 & 0.098 & 0.156 & 0.084 & 0.050 \\
         & &\cellcolor{graybox}{SELF \cite{MOSTEGEL_CVPR_2016}} & 0.128 & 0.105 & 0.117 & 0.066 & 0.091 & 0.154 & 0.086 & 0.060 \\  
         & &\cellcolor{blackbox}\bfseries {\otb{} (Ours)} & \bfseries0.127 & \bfseries\leavevmode\color{red}0.084 & \bfseries\leavevmode\color{red}0.111 & \bfseries\leavevmode\color{red}0.056 & \bfseries0.090 & \bfseries0.139 & \bfseries\leavevmode\color{red}0.064 & \bfseries\color{red}0.035 \\

        \hline\hline 
         & & Optimal & 0.053 & 0.041 & 0.046 & 0.022 & 0.057 & 0.103 & 0.030 & 0.014 \\    
        \hline
    \end{tabular}
    }
        \caption{\textbf{Generalization on Middlebury and ETH3D.} We report AUC scores for networks trained on KITTI 2012 (20 or 400 images) and tested on Middlebury (15 images) and ETH3D (27 images) without retraining or adaptation.}
    \label{tab:middeth}
\end{table}

\begin{figure*}[t]
    \centering
    \renewcommand{\tabcolsep}{1pt}
    \begin{tabular}{cccccc}
        \includegraphics[width=0.16\textwidth]{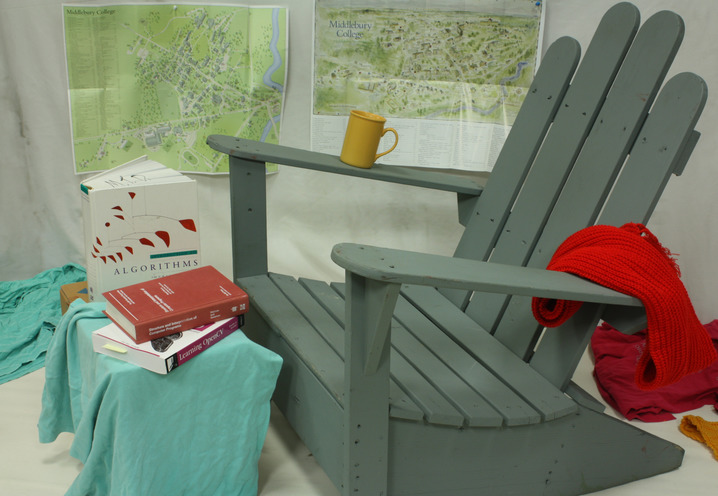} &
        \includegraphics[width=0.16\textwidth]{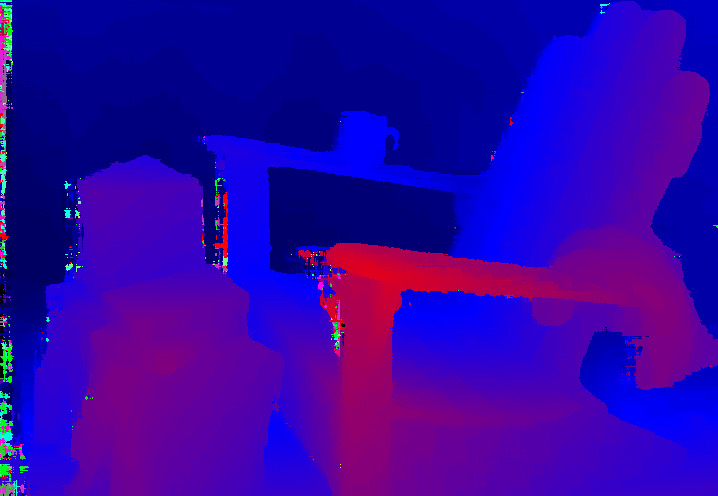} &
        \includegraphics[width=0.16\textwidth]{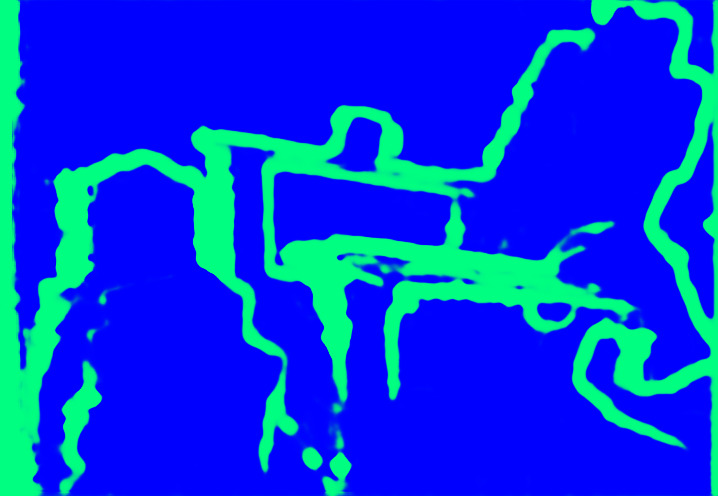} &
        \includegraphics[width=0.16\textwidth]{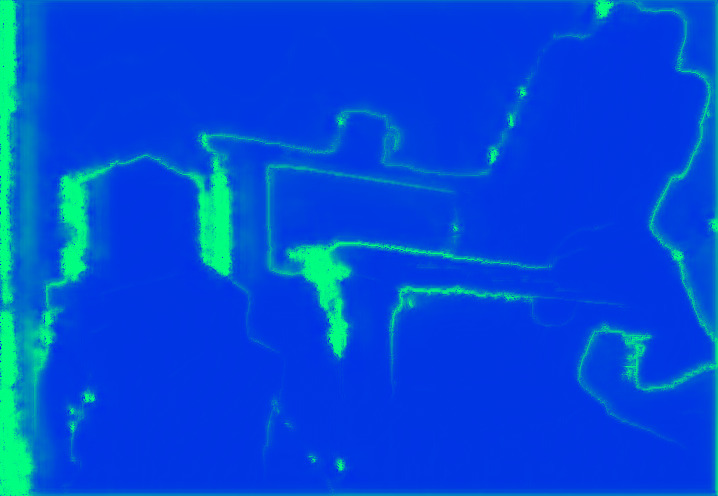} &
        \includegraphics[width=0.16\textwidth]{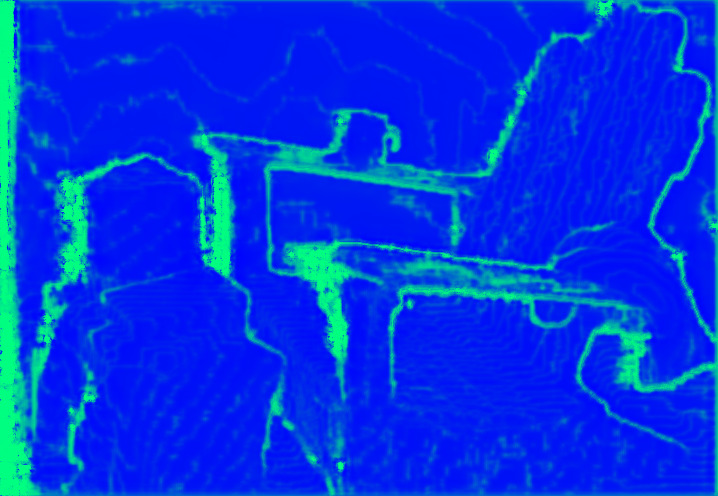}
        \includegraphics[width=0.16\textwidth]{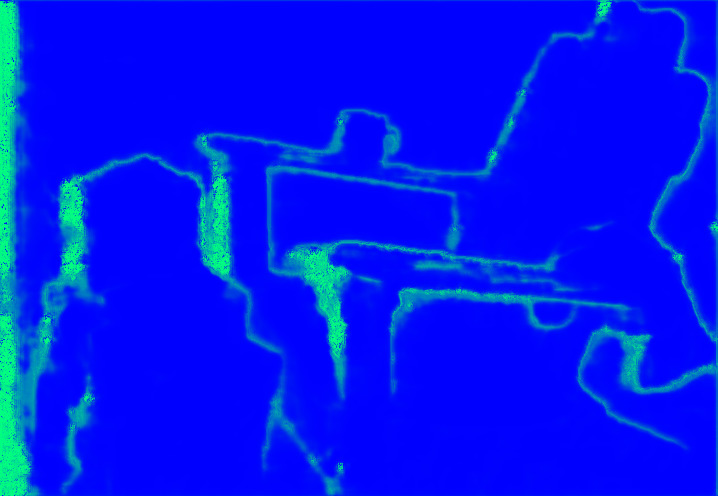}\\
        
        \includegraphics[width=0.16\textwidth]{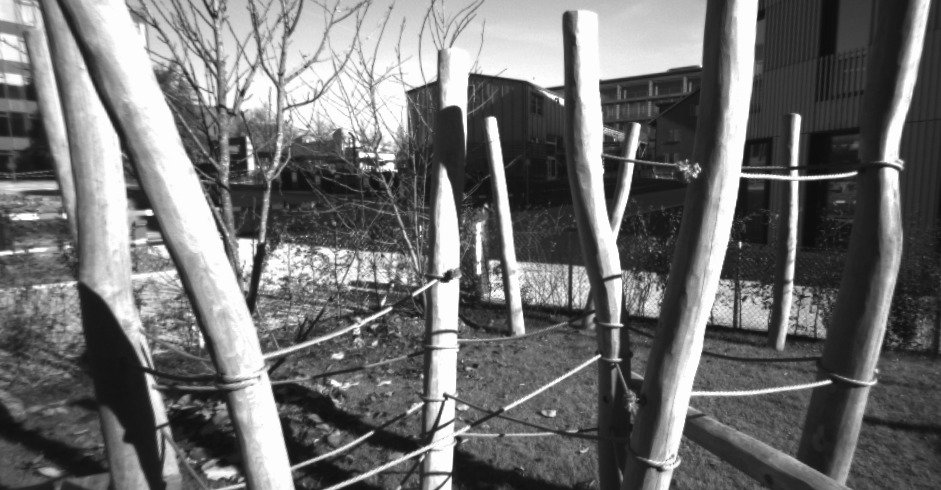} &
        \includegraphics[width=0.16\textwidth]{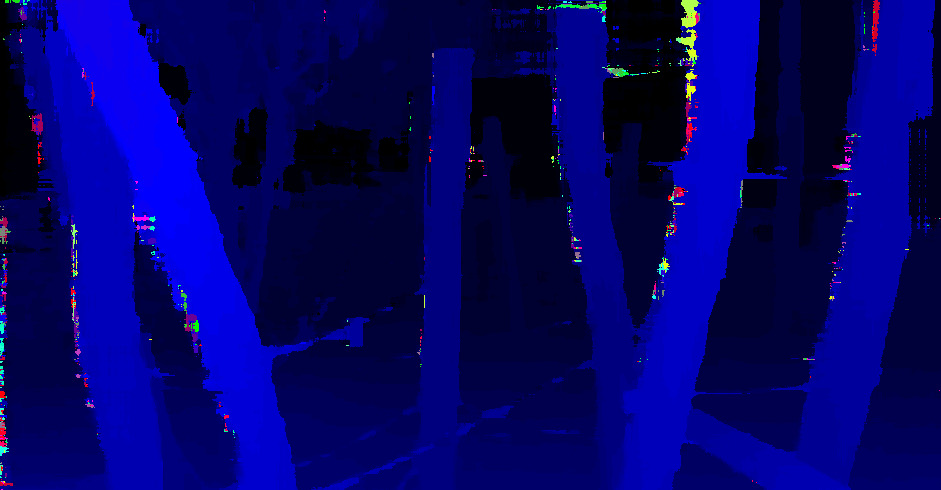} &
        \includegraphics[width=0.16\textwidth]{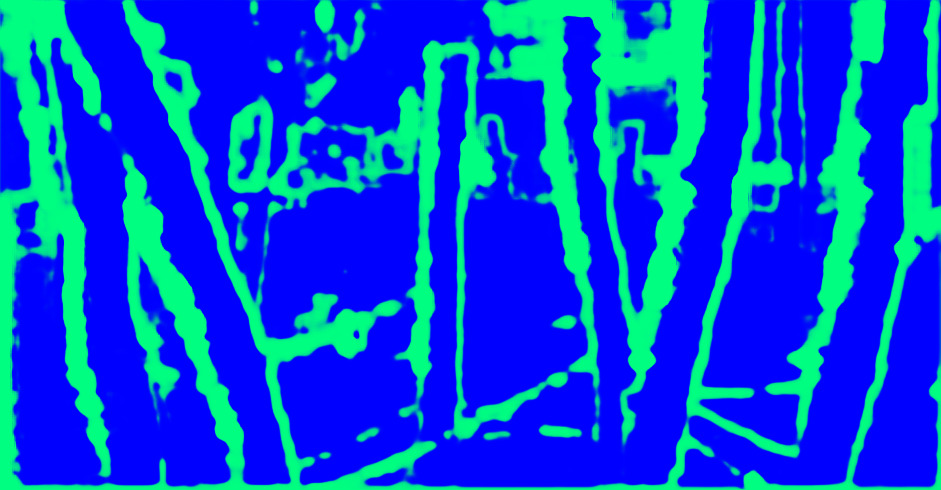} &
        \includegraphics[width=0.16\textwidth]{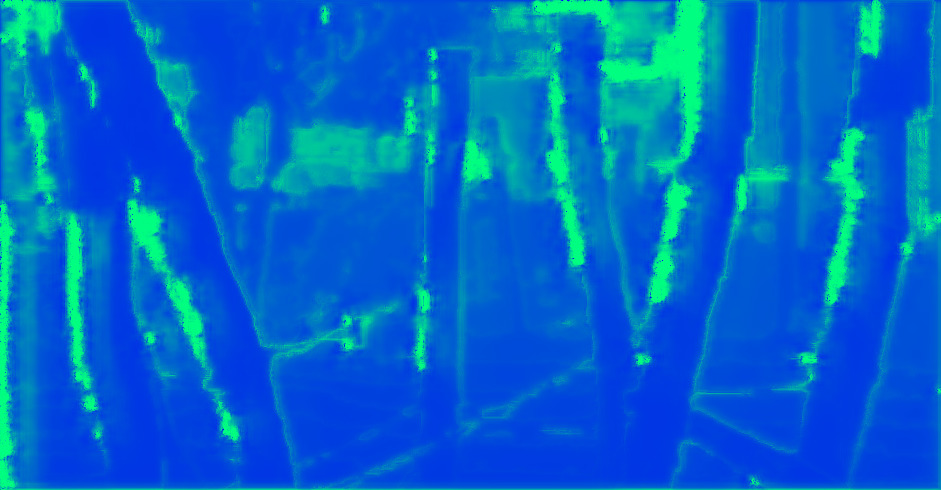} &
        \includegraphics[width=0.16\textwidth]{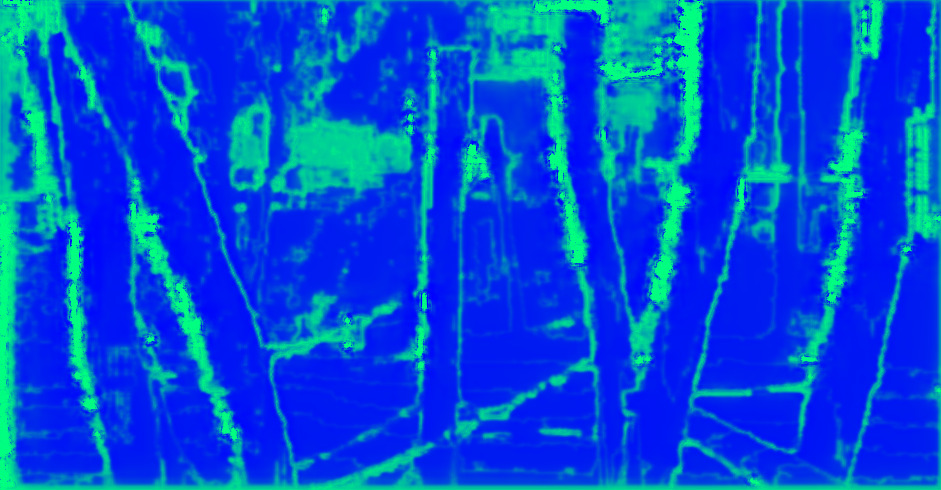}
        \includegraphics[width=0.16\textwidth]{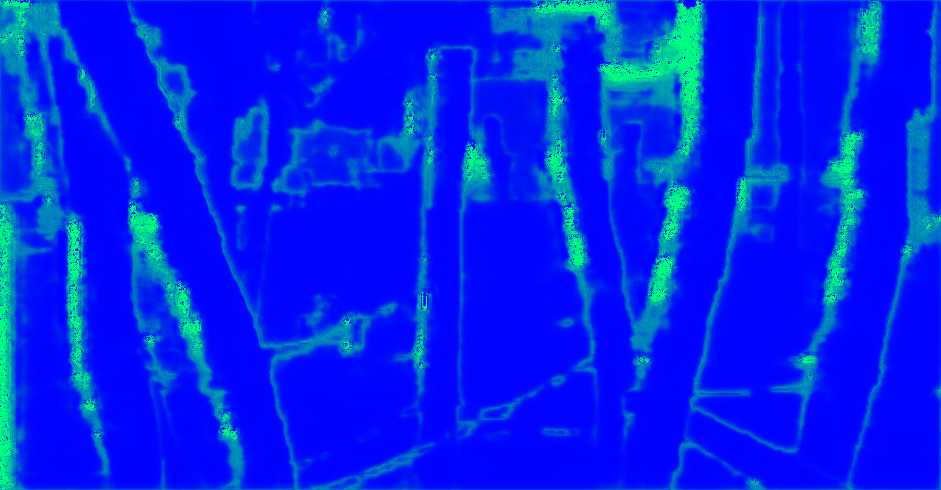}\\
    \end{tabular}
    
    \caption{\textbf{Qualitative results for generalization.} From left: reference image, disparity by MCCNN-fst-SGM, ConfNet trained with \cite{Tosi_2017_BMVC}, \cite{MOSTEGEL_CVPR_2016}, our method and ground-truth. On top, Adirondack (Middlebury), at the bottom, Playground\_3l (ETH3D).}
    \label{fig:middeth}
\end{figure*}

\textbf{KITTI datasets.} Table \ref{tab:kitti} collects evaluations on the KITTI 2012 and 2015 datasets, respectively, using the 174 validation set from 2012 and the full 2015 set. 
We point out that all self-supervised strategies outperform traditional measures, reported on top as baselines, such as LRD, PKR, LRC and the cues used in our $\mbce$ loss, struggling only when dealing with the very accurate MCCNN-fst-SGM algorithm. Comparing the different architectures, we can notice how the self-supervised paradigms break the hierarchy (\ie, self-supervised LGC is often outperformed by ConfNet). On Census-CBCA, our strategy always outperforms SELF and WILD when used to train any architecture. The same behaviour is confirmed on MCCNN-fst-CBCA, except for CCNN resulting better with SELF but with the best performance achieved by ConfNet trained with \otb{}. This outcome highlights the outstanding performance of \otb{} with noisy algorithms (about 27 and 19\% error rates on the validation set), close to \textcolor{red}{\textbf{full supervision}}. On SGM algorithms, \otb{} results comparable with SELF and WILD, although sourcing supervision only from images and $\dispL$, thus in a much weaker form compared to the competitors. On three out of four algorithms, ConfNet results to be the most effective architecture when trained in a self-supervised manner.

\textbf{Generalization on Middlebury and ETH3D.} Table \ref{tab:middeth} reports results on the Middlebury 2014 and ETH3D datasets. The same networks evaluated so far (trained on KITTI 2012 images) are transferred here without retraining or adaptation, enabling to assess the generalization properties of each network/supervision configuration. Not surprisingly, the margin between learned and traditional measures is much smaller because of the domain shift. Nonetheless, in many cases the performance is still in favor of learned approaches, with some exceptions. We point out that networks trained with \otb{} self-supervision always outperform SELF and WILD, except for ConfNet in two cases. Moreover, networks trained with \otb{} generalize better than their fully supervised counterparts in some cases, mostly on the ETH3D dataset (\eg, CCNN with all algorithms, ConfNet with MCCNN-fst-SGM and LGC with both SGM methods). 
Fig. \ref{fig:middeth} shows qualitative examples of this test. 

\begin{table}[t]
    \centering
    \scalebox{0.72}{
    \renewcommand{\tabcolsep}{3pt}
    \begin{tabular}{l | ccc| c| cccc | cc }
         \multicolumn{1}{c}{} & \multicolumn{3}{c}{Traditional} & \multicolumn{5}{c}{Supervision} & & \\
         \hline
         Algorithm & \cellcolor{blackbox}$\Delta_{(\imageL,\Tilde{\imageR})}$ & \cellcolor{blackbox}DA & \cellcolor{blackbox}UC & \cellcolor{supervised}Supervised &  \cellcolor{whitebox}WILD \cite{Tosi_2017_BMVC} & \cellcolor{graybox}SELF \cite{MOSTEGEL_CVPR_2016} & \cellcolor{blackbox}\otb{} & \cellcolor{blackbox}\textbf{\otb{} (online)} & Opt. & Bad$\tau$ \%\\
        \hline
        Census-SGM & 0.179 & 0.106 & 0.162 & 0.061  & 0.067 & 0.074 & 0.072 & \bfseries\leavevmode\color{red}0.064 & 0.029 & 21.007 \\
        \hline
        MADNet \cite{Tonioni_2019_CVPR} & 0.133 & 0.147 & 0.155 & 0.116 & - & 0.135 & 0.139 & \bfseries\leavevmode\color{red}0.126 & 0.021 & 16.226\\
        \hline
        \hline
        GANet \cite{Zhang2019GANet} & 0.019 & 0.018 & 0.025 & 0.017 & - & 0.021 & 0.019 & \bfseries\leavevmode\color{red}0.015 & 0.001 & 2.897\\
        \hline\hline
        
    \end{tabular}
    }
        \caption{\textbf{Self-adaptation.} We report AUC scores for networks trained on KITTI 2012 (20 or 400 images) and tested on a DrivingStereo sequence (6905 frames).}
    \label{tab:adaptation}
\end{table}

\begin{figure}[t]
    \centering
    \renewcommand{\tabcolsep}{1pt}
    \begin{tabular}{cccccc}
        \includegraphics[width=0.16\textwidth]{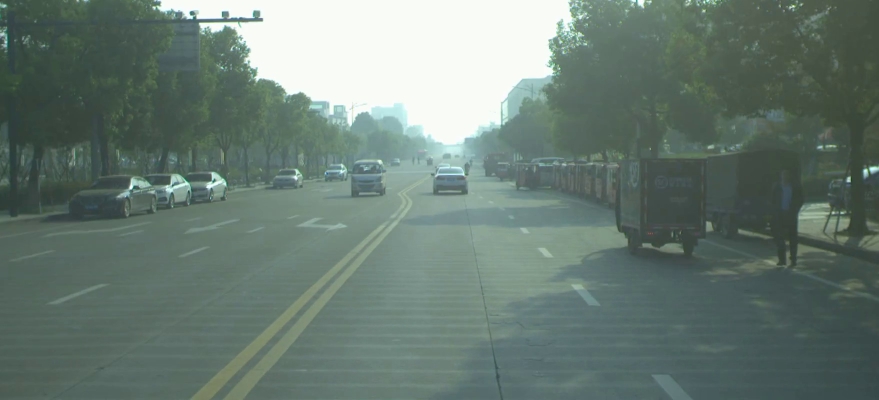} &
        \includegraphics[width=0.16\textwidth]{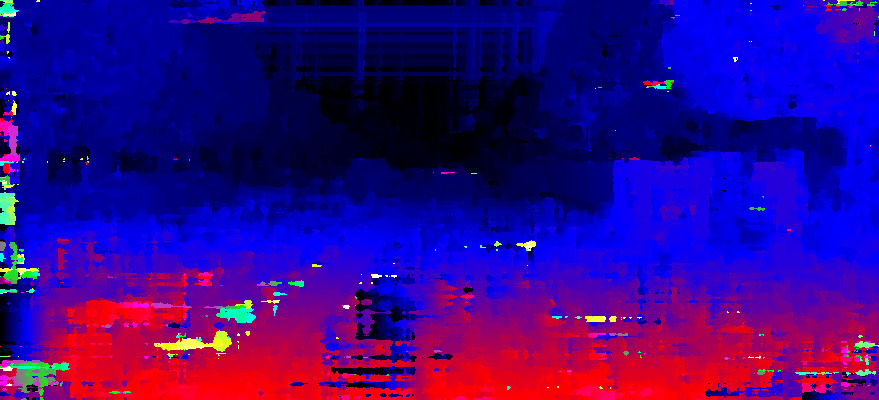} &
        \includegraphics[width=0.16\textwidth]{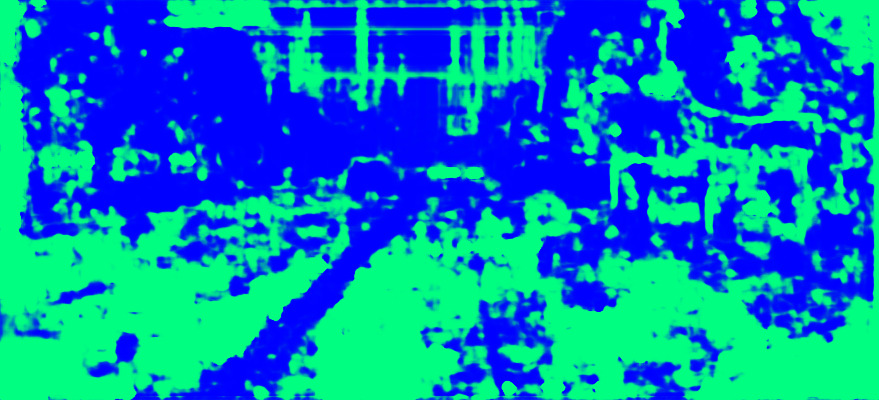} &
        \includegraphics[width=0.16\textwidth]{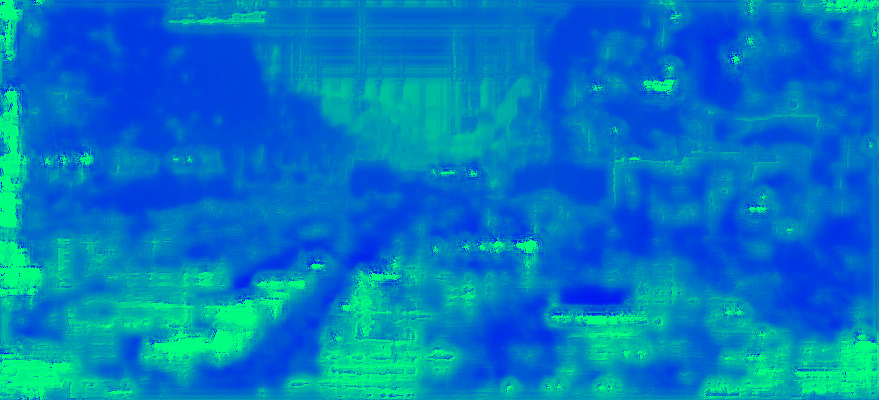} &
        \includegraphics[width=0.16\textwidth]{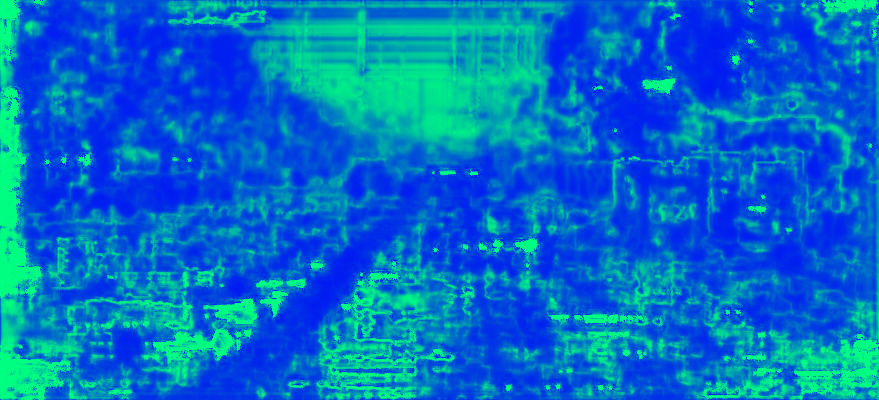}
        \includegraphics[width=0.16\textwidth]{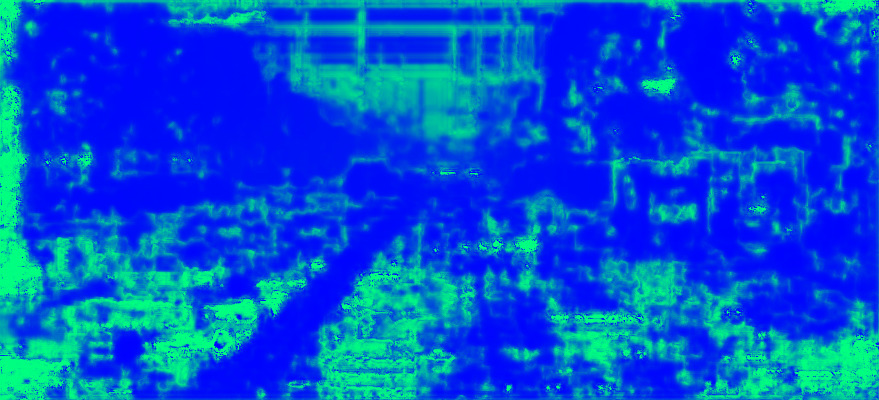} \\

    \end{tabular}
    
    \caption{\textbf{Qualitative results on DrivingStereo.} From left: reference image, disparity by Census-SGM, ConfNet trained with \cite{Tosi_2017_BMVC}, \cite{MOSTEGEL_CVPR_2016}, \otb{} and online-adapted \otb{}.}
    \label{fig:drivingstereo}
\end{figure}

\subsection{Self-adapting in-the-wild}

Finally, we conduct experiments aimed at assessing how effective our strategy is for self-adaptation of a confidence measure in unseen environments. Purposely, we simulate deployment in an autonomous driving scenario, selecting a sequence from the DrivingStereo dataset \cite{yang2019drivingstereo}. We use sequence 2018-10-25-07-37, containing 6905 stereo pairs acquired in unconstrained (\ie, dynamic) environment. For this evaluation, we choose Census-SGM, MADNet and GANet. The former because it represents the preferred choice for hardware implementation on custom stereo cameras \cite{HW_SGM_BANZ,HW_SGM_MERCEDES,Oberpfaffenhofen,POLLEFEYS_SGM,smartcamera,rahnama2018r3sgm,rahnama2019real}. The remaining two because well representing modern end-to-end CNNs that are fast (MADNet) or yield state-of-the-art accuracy (GANet).
For confidence networks we select ConfNet, because effective with accurate algorithms and well-suited for online adaptation.

In this experiment, we assume to have pre-trained versions of ConfNet with the different self-supervision paradigms, again on KITTI 2012. For \otb{}, we use $[\texture^p,\agreement^p,\uniqueness^p,\texture^q,\agreement^q,\uniqueness^q]$ for offline training and $[\texture^p,\agreement^p,\uniqueness^p,\texture^q]$ during adaptation.
When performing online adaptation (\textbf{online} entry), for each stereo pair the confidence is estimated and evaluated \textbf{before} loss computation (thus, supervision only acts on the upcoming frames as in \cite{Tonioni_2019_CVPR}). This way, ConfNet runs at 0.09 seconds (11 FPS) against the 0.02 (50 FPS) without adaptation on Titan Xp, measuring network computations only. The learning rate is set to 0.0001 during adaptation.
Table \ref{tab:adaptation} collects the outcome of this evaluation. 
We point out that WILD can not be deployed for MADNet and GANet since a meaningful cost volume is not available for the former or cannot be used straightforwardly for the latter. On the other hand, SELF would require ($\dispL,\dispR)$ for supervision, while MADNet and GANet provide only the former. By assuming networks as gray-boxes, we get rid of this issue at training time obtaining $\dispR$ as shown in Eq. \ref{eq:flip}. Concerning SGM, \otb{} performs in between WILD and SELF. Nevertheless, keeping continuous adaptation active on the whole sequence makes it outperform both. Concerning MADNet, SELF results more effective than \otb{}. Again, performing online adaptation makes \otb{} the best solution in this case as well. Finally, concerning GANet, learned measures perform worse than $\Delta_{(\imageL,\Tilde{\imageR})}$ and DA. Online adaptation results crucial for \otb{} to obtain the best results. 
To conclude, Fig. \ref{fig:drivingstereo} shows qualitative examples for the SGM algorithm.

\begin{figure}[t]
    \centering
    \begin{tabular}{cccccc}
        \includegraphics[width=0.14\textwidth]{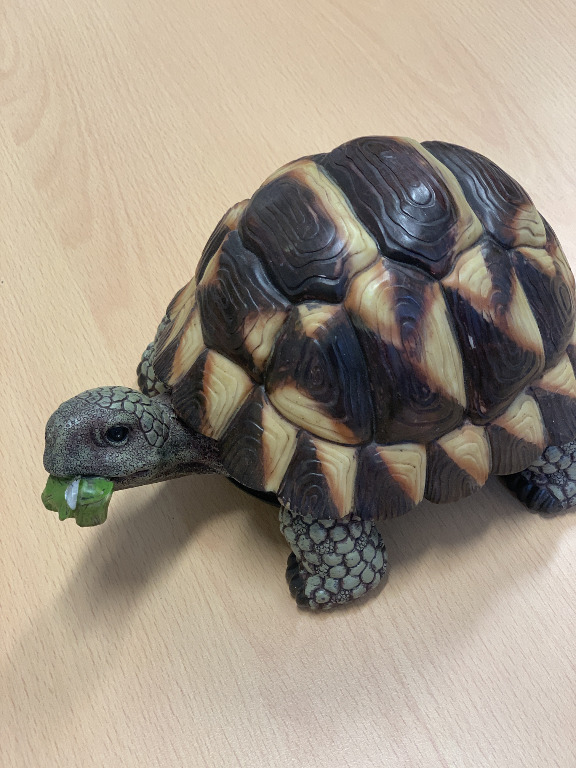} &
        \includegraphics[width=0.14\textwidth]{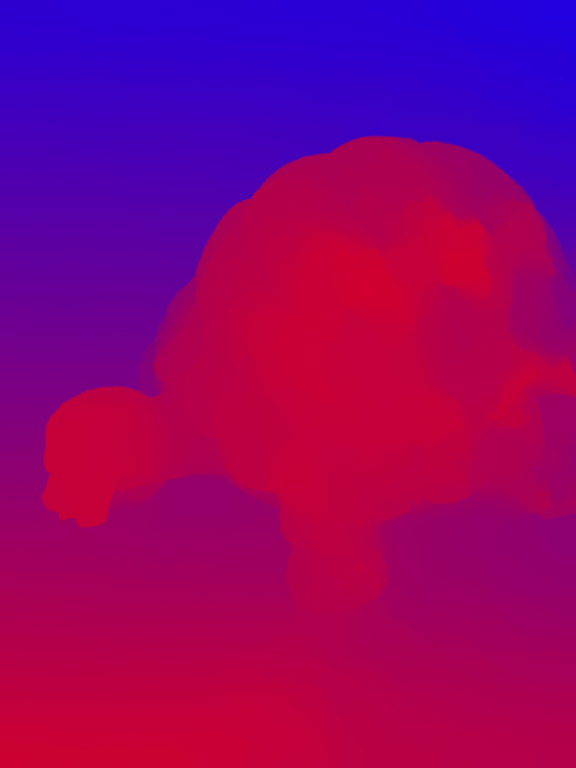} &
        \includegraphics[width=0.14\textwidth]{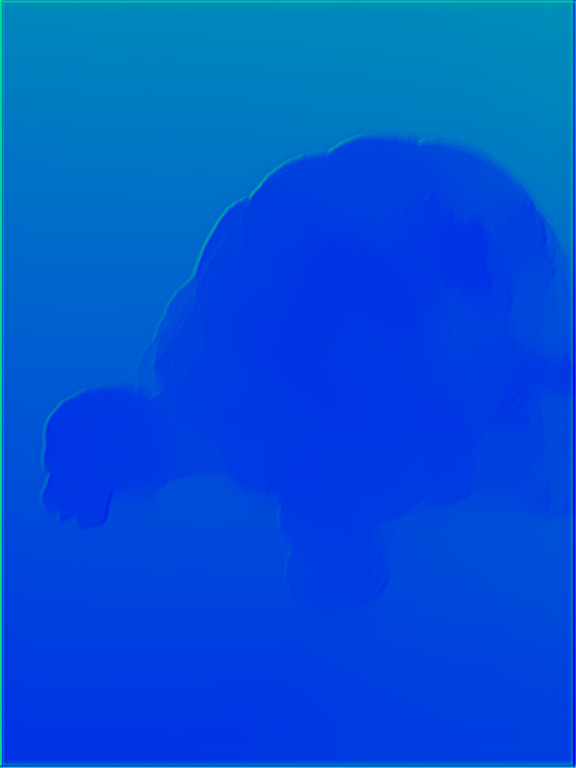} &
        \includegraphics[width=0.14\textwidth]{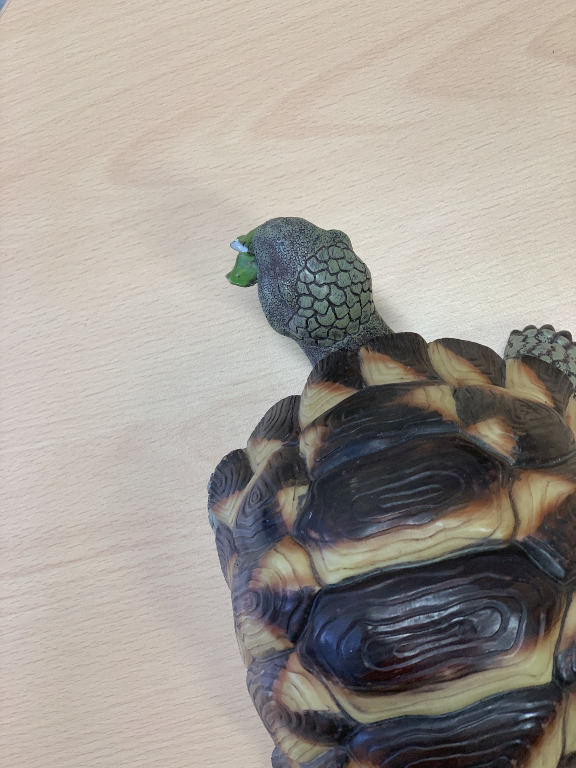} &
        \includegraphics[width=0.14\textwidth]{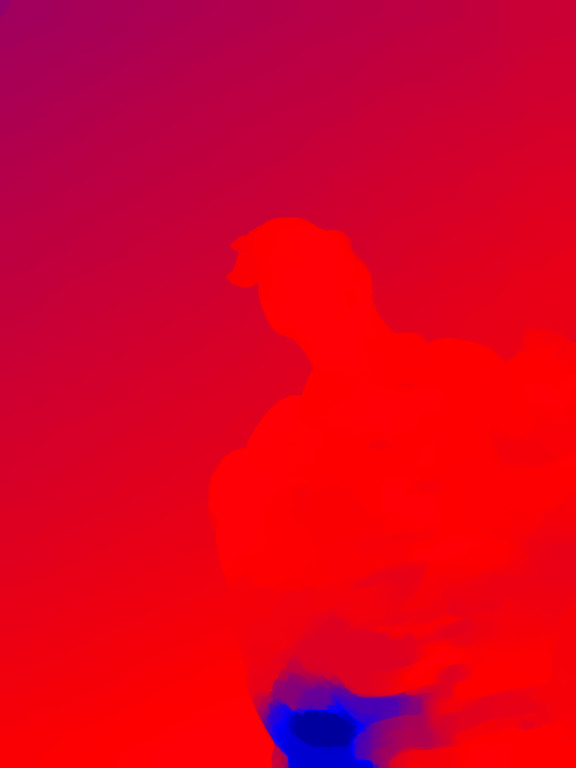} &
        \includegraphics[width=0.14\textwidth]{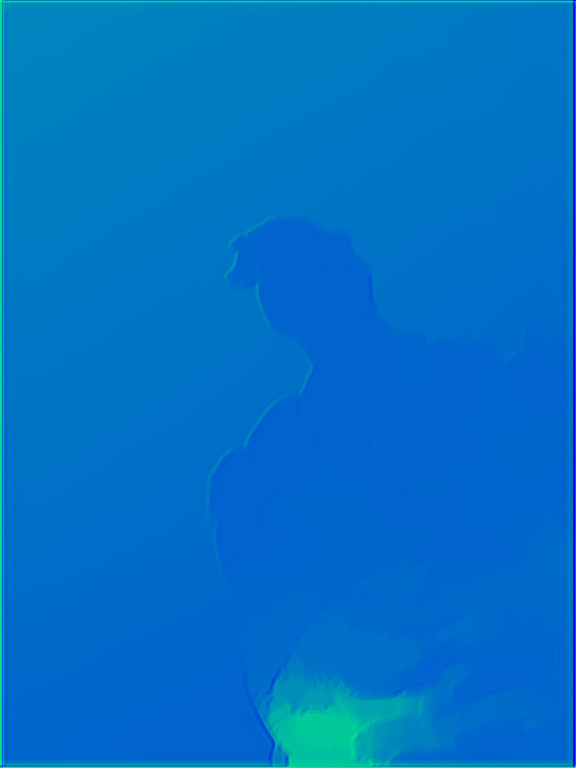} \\        
    \end{tabular}
    \caption{\textbf{Qualitative results with Apple iPhone XS.} We show two examples of reference image and disparity map acquired with the iPhone XS, followed by estimated confidence map after few iterations of on-the-fly learning.}
    \label{fig:iphone}
\end{figure}

\textbf{On-the-fly learning with black-box sensors.} Finally we report, as qualitative results, the outcome obtained by learning on-the-fly a confidence measure on disparity map sourced by an Apple iPhone XS, without any pre-training. Fig. \ref{fig:iphone} shows examples of acquired disparity and estimated confidence maps by ConfNet adapted online, detecting gross errors like on turtle's shell.

\section{Conclusion}

In this paper, we have introduced a novel self-supervised paradigm aimed at learning from scratch a confidence measure for stereo. We leverage few, principled cues from the input stereo pair and the estimated disparity in order to source supervision signals in place of disparity ground truth labels. Being such cues available during deployment in-the-wild, our solution is suited for continuous online adaptation on any black-box framework. Experimental results proved that our strategy is equivalent or superior to existing self-supervised approaches and, conversely to them, allows for further improvements during deployment by leveraging the online self-adaptation process.

\textbf{Acknowledgments.} We gratefully acknowledge the support of NVIDIA Corporation with the donation of the Titan Xp GPU used for this research.

\clearpage

\bibliographystyle{splncs04}
\bibliography{egbib}
\end{document}